\documentclass[conference]{IEEEtran}
\IEEEoverridecommandlockouts

\usepackage{times}
\usepackage{amsmath}
\usepackage{amssymb}
\usepackage[sort&compress, numbers]{natbib}
\usepackage{url}
\usepackage{xcolor}
\usepackage{fancyhdr}
\usepackage{adjustbox}
\usepackage{graphicx}
\usepackage{subcaption} %
\usepackage{tensor}
\usepackage{tabularx}
\usepackage{multirow}
\usepackage{multicol}
\usepackage{colortbl}
\usepackage[utf8]{inputenc}
\usepackage[OT1]{fontenc}
\usepackage[final]{microtype}
\usepackage{booktabs}
\usepackage[binary-units]{siunitx}
\usepackage{listings}
\usepackage{placeins}

\usepackage[font=small,labelfont=bf]{caption}
\usepackage{cuted}
\usepackage[abbreviations]{glossaries-extra}

\glssetcategoryattribute{abbreviation}{indexonlyfirst}{true}

\glssetcategoryattribute{abbreviation}{nohyperfirst}{true}

\newabbreviation{auroc}{AUROC}{Area Under the Receiver Operating Characteristic Curve}
\newabbreviation{accuracy}{Acc}{Accuracy}

\newabbreviation{cnn}{CNN}{Convolutional Neural Network}

\newabbreviation{fov}{FoV}{Field of View}
\newabbreviation{fpr}{FPR}{False Positive Ratio}
\newabbreviation{fem}{FEM}{Finite Element Method}
\newabbreviation{fdm}{FDM}{Forward Dynamics Mode}

\newabbreviation{pomdp}{POMDP}{Partially Observable Markov Decision Process}

\newabbreviation{gnn}{GNN}{Graph Neural Network}
\newabbreviation{gcn}{GCN}{Graph Convolutional Network}
\newabbreviation{imu}{IMU}{Inertial Measurement Unit}
\newabbreviation{irl}{IRL}{Inverse Reinforcement Learning}

\newabbreviation{knn}{KNN}{K-Nearest Neighbors}

\newabbreviation{lagr}{LAGR}{Learning Applied to Ground Vehicles}
\newabbreviation{lio}{LIO}{LiDAR Interial Odometry}

\newabbreviation{mlp}{MLP}{Multi-Layer Perceptron}
\newabbreviation{mpc}{MPC}{Model Predictive Controller}
\newabbreviation{mppi}{MPPI}{Model Predictive Path Integral}
\newabbreviation{mse}{MSE}{Mean Squared Error}
\newabbreviation{mdp}{MDP}{Markov Decision Process}

\newabbreviation{ood}{OOD}{out-of-distribution}

\newabbreviation{rbf}{RBF}{Radial Basis Function}
\newabbreviation{rmp}{RMP}{Riemannian Motion Policies}
\newabbreviation{ros}{ROS}{Robot Operating System}
\newabbreviation{ros1}{ROS~1}{Robot Operating System}
\newabbreviation{roc}{ROC}{Receiver Operating Characteristic}
\newabbreviation{rf}{RF}{Random Forest}

\newabbreviation{sdf}{SDF}{Signed Distance Field}
\newabbreviation{slam}{SLAM}{Simultaneous Localization and Mapping}
\newabbreviation{svm}{SVM}{Support Vector Machine}
\newabbreviation{svc}{SVC}{Support Vector Classifier}
\newabbreviation{wvn}{WVN}{Wild Visual Navigation}

\newabbreviation{vit}{ViT}{Vision Transformer}
\newabbreviation{vio}{VIO}{Visual Interial Odometry}

\newcommand\mytxcellwidth{\TX@col@width}

\usepackage[
    hidelinks,
    pdfauthor={Pascal Roth},
    pdftitle={Learned Perceptive Forward Dynamics Model for Safe and Platform-aware Robotic Navigation},
    pdfsubject={Robots},
    pdfkeywords={Robots, Learning, Navigation, Dynamics Model}
]{hyperref}
\hypersetup{
    colorlinks=true,     %
    linkcolor=blue,      %
    urlcolor=blue,       %
    citecolor=blue,      %
}

\newcolumntype{Y}{>{\centering\arraybackslash}X}

\usepackage{misc/onimage}

\tikzset{
    image label/.style={
        every node/.style={
            fill=black,
            text=white,
            font=\fontfamily{phv}\selectfont\scriptsize\bfseries,
            anchor=north west,
            xshift=0.05cm,
            yshift=-0.15cm,
            at={(0,1)}
        }
    }
}

\tikzset{
    image label_tl/.style={
        every node/.style={
            fill=black,
            text=white,
            font=\fontfamily{phv}\selectfont\scriptsize\bfseries,
            anchor=north west,
            xshift=0.1cm,
            yshift=-0.1cm,
            at={(0,1)}
        }
    }
}

\tikzset{
    image label_tr/.style={
        every node/.style={
            fill=black,
            text=white,
            font=\fontfamily{phv}\selectfont\scriptsize\bfseries,
            anchor=north east,
            xshift=-0.1cm,
            yshift=-0.1cm,
            at={(1,1)}
        }
    }
}

\tikzset{
    image label_bl/.style={
        every node/.style={
            fill=black,
            text=white,
            font=\fontfamily{phv}\selectfont\scriptsize\bfseries,
            anchor=south west,
            xshift=0.1cm,
            yshift=0.1cm,
            at={(0,0)}
        }
    }
}

\tikzset{
    image label_br/.style={
        every node/.style={
            fill=black,
            text=white,
            font=\fontfamily{phv}\selectfont\scriptsize\bfseries,
            anchor=south east,
            xshift=-0.1cm,
            yshift=0.1cm,
            at={(1,0)}
        }
    }
}

\lstdefinestyle{python}{
    language=Python,
    backgroundcolor=\color{backcolour},   
    commentstyle=\color{codered}\textit,
    keywordstyle=\bfseries\color{codegreen},
    numberstyle=\tiny\color{codegray},
    stringstyle=\color{codepurple},
    basicstyle=\ttfamily\tiny,  %
    breakatwhitespace=false,         
    breaklines=true,                 
    captionpos=b,                    
    keepspaces=true,                 
    numbers=left,                    
    numbersep=4pt,                  
    showspaces=false,                
    showstringspaces=false,
    showtabs=false,                  
    tabsize=1,
    fancyvrb=true
}

\definecolor{ours}{rgb}{0.3686274509803922, 0.5058823529411764, 0.6745098039215687}
\definecolor{baseline}{rgb}{0.6549019607843137, 0.8, 0.43137254901960786}
\definecolor{constant_vel}{rgb}{0.5882352941176471, 0.1411764705882353, 0.5686274509803921}
\definecolor{trajectory}{rgb}{0.984313725490196, 0.592156862745098, 0.15294117647058825}
\definecolor{collision}{rgb}{0.823529412, 0.168627451, 0.149019608}

\title{
    \vspace{1pt}
    \LARGE \bf 
    Learned Perceptive Forward Dynamics Model for Safe and Platform-aware Robotic Navigation
    \vspace{-8pt}
}

\author{\authorblockN{Pascal Roth\authorrefmark{1}\authorrefmark{2},
Jonas Frey\authorrefmark{1}\authorrefmark{3},
Cesar Cadena\authorrefmark{1}, and
Marco Hutter\authorrefmark{1}%
}
\authorblockA{\authorrefmark{1}
    ETH Zurich  \authorrefmark{2} NVIDIA \authorrefmark{3} Max Planck Institute for Intelligent Systems\\ 
    \{rothpa, jonfrey, cesarc, mahutter\}@ethz.ch
}}

\usepackage{fancyhdr}
\newcommand{\mytitle}{\textbf{Accepted version.} To appear in the proceedings of \textit{RSS 2025}} 
\begin{document}

\makeatletter
    \let\@oldmaketitle\@maketitle%
    \renewcommand{\@maketitle}{
    \@oldmaketitle
    \centering
    \vspace{-2.5pt}
    \includegraphics[width=\textwidth]{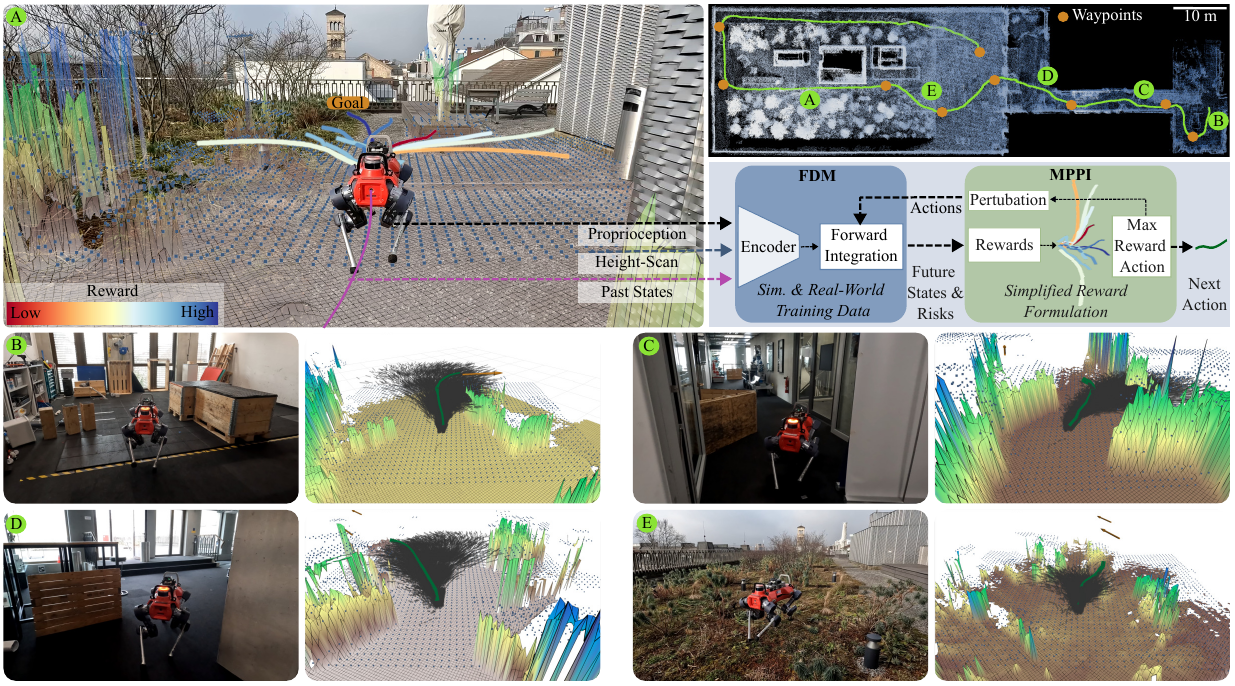}
    \captionof{figure}{
    Demonstration of the proposed perceptive Forward Dynamics Model for robust navigation in complex environments. The model, trained with real-world and simulation data, predicts the robot’s future states given a sequence of velocity actions. It takes as input the surrounding geometry in the form of a height scan, along with past states and proprioceptive measurements. A sampling-based planner evaluates the integrated paths based on simple reward functions to select the optimal next action in a receding horizon fashion.
    (A) Ten example paths are visualized and overlaid on the environment image alongside the height map and the downsampled height scan (blue points). Path colors indicate rewards, with the highest reward assigned to the closest collision-free trajectory to the goal.
    (B–E) Additional planning events are shown, displaying sampled paths and the selected trajectory (green), demonstrating safe planning in rough terrain.}
    \label{fig:real_world_planning}
    \vspace{-9.5pt}
    \addtocounter{figure}{-1}%
    }
\makeatother

\maketitle

\addtocounter{figure}{1}%

\thispagestyle{fancy}

\begin{abstract}
Ensuring safe navigation in complex environments requires accurate real-time traversability assessment and understanding of environmental interactions relative to the robot's capabilities. Traditional methods, which assume simplified dynamics, often require designing and tuning cost functions to safely guide paths or actions toward the goal. This process is tedious, environment-dependent, and not generalizable. To overcome these issues, we propose a novel learned perceptive Forward Dynamics Model (FDM) that predicts the robot’s future state conditioned on the surrounding geometry and history of proprioceptive measurements, proposing a more scalable, safer, and heuristic-free solution. The FDM is trained on multiple years of simulated navigation experience, including high-risk maneuvers, and real-world interactions to incorporate the full system dynamics beyond rigid body simulation. We integrate our perceptive FDM into a zero-shot Model Predictive Path Integral (MPPI) planning framework, leveraging the learned mapping between actions, future states, and failure probability. This allows for optimizing a simplified cost function, eliminating the need for extensive cost-tuning to ensure safety. On the legged robot ANYmal, the proposed perceptive FDM improves the position estimation by on average 41\% over competitive baselines, which translates into a 27\% higher navigation success rate in rough simulation environments.  Moreover, we demonstrate effective sim-to-real transfer and showcase the benefit of training on synthetic and real data. Code and models are made publicly available under \url{https://github.com/leggedrobotics/fdm}.
\end{abstract}

\IEEEpeerreviewmaketitle

\section{Introduction}

Understanding robotic system dynamics is essential for ensuring safe and effective control, particularly in complex tasks like motion planning in contact-rich scenarios. The dynamics of a mobile robot navigating within an environment depend on its structure and interactions with the terrain. This results in highly nonlinear behaviors that are challenging to generalize across diverse scenarios~\cite{gibson2023multi}.
\textit{Forward Dynamics Models} (FDM) are typically used to predict such complex dynamics, estimating the robot's future state conditioned on the applied commands.
These models capture the robot-terrain interactions and implicitly provide a terrain traversability estimate.
Dynamic models must carefully balance key modeling choices, including state representation, fidelity, time horizon, and modeling frequency. 
While the dynamics of on- and off-road vehicles have been extensively explored~\cite{xiao2021learning, pokhrel2024cahsor, gibson2024dynamics}, quadrupedal robots present unique challenges due to their complex system dynamics and difficult-to-model environmental interactions~\cite{kim2022learning}. 
Moreover, their learned locomotion policies, which rely on deep neural networks, additionally complicate the modeling of the robot's behavior. 

Traditional physics-based models that are derived from first principles and calibrated using system identification often fail to capture the dynamics accurately. They specifically struggle in contact-rich scenarios, which introduce additional non-linearities and require accurate perception~\cite{Duong2024PortHamiltonianNO}. Further, they can be computationally expensive and sensitive to initial conditions. Consequently, those models face challenges when it comes to accurately modeling system dynamics, which in turn results in biased predictions and persistent modeling errors.

To overcome these limitations, data-driven approaches have emerged as a promising alternative to approximate complex dynamics. However, training neural networks to represent robot dynamics often requires substantial amounts of state-action trajectories, motivating the use of synthetic data to mitigate the challenges of collecting extensive real-world datasets~\cite{kim2022learning}. Further, simulation allows performing dangerous or catastrophic maneuvers that harm the real robot, such as falling or colliding. 
While the simulator's complex physics modeling is accurate in rigid-body scenarios, it is computationally expensive and fails to capture scenarios outside of its domain. As a result, it becomes necessary to distill the dynamics into a learned model for sufficient inference speed on a compute-restricted mobile robot~\cite{kim2022learning}. Moreover, real-world data remains essential for addressing unmodeled effects and bridging the reality gap~\cite{kahn2021badgr}.
Addressing the gap between physics-informed and learned models, approaches that integrate physics constraints — such as kinematic laws or energy conservation — in the learning setup show strong performance~\cite{lutter2023combining, cranmer2020lagrangian, roehrl2020modeling, Duong2024PortHamiltonianNO}. However, they remain limited to short control timescales compared to the longer planning timesteps addressed in this work.
The first work that employs a learned FDM on a quadrupedal system has been done by~\cite {kim2022learning}. Combined with their developed trajectory sampling technique, they demonstrate reactive navigation in complex, narrow environments. However, open challenges remain to incorporate $3D$ perception to target rough environments and the transfer from simulation to the real system.

This work introduces a perceptive, $n$-step Forward Dynamics Model framework. The proposed approach combines pre-training with synthetic data generated using a state-of-the-art simulator and fine-tuning with real-world data. This hybrid strategy leverages the safety and flexibility of simulation while capturing real-world dynamics. The FDM is designed for both legged and wheel-legged systems, marking the first application of its kind in rough terrain environments.
Our novel framework extends the capabilities of sampling-based planner methods by reducing the need for extensive parameter tuning and providing a flexible solution for non-task-specific planning. This enables zero-shot adaptation to new environments without requiring additional learning steps. Additionally, the perception capabilities of the model represent a significant step forward, offering an attractive alternative to explicitly modeling the environment's traversability. 
The main contributions of this work are as follows:

\begin{itemize}
    \item The first application of a rough-terrain Forward Dynamics Model trained in simulation and deployed on a quadrupedal robot. The model demonstrates reliable sim-to-real transfer capabilities and robust performance in rough terrain.
    \item A hybrid training strategy using real-world data to effectively capture the full system's dynamics beyond rigid-body simulation while leveraging synthetic data for pre-training to safely account for high-risk scenarios.
    \item A simplified cost formulation for MPPI-based planning that integrates the platform-specific FDM to enable safe and reliable trajectory generation. The approach supports zero-shot adaptation to new environments by cost-term adjustments without the need for additional training.
\end{itemize}

\section{Related Work}

\subsection{Dynamics Modeling}

The field of dynamics learning has predominantly focused on data-driven solutions~\cite{xiao2021learning, pokhrel2024cahsor, kim2023bridging, kahn2021badgr, lee2023learning, gibson2024dynamics, gibson2023multi}, as models derived from first principles and calibrated via system identification often oversimplify or misrepresent system dynamics, leading to bias and persistent modeling errors~\cite{Duong2024PortHamiltonianNO}. In contrast, learned models can approximate complex, nonlinear dynamics from large datasets~\cite{deisenroth2011pilco, levine2014learning} while capturing uncertainties using probabilistic neural network ensembles~\cite {chua2018deep, kim2023bridging}. 
To incorporate environmental context, approaches integrate terrain information via geometric measurements such as 2D LiDAR scans~\cite{kim2022learning} or height maps~\cite{lee2023learning, gibson2024dynamics, fabian2020pose}, as well as RGB images~\cite{kahn2021badgr, bar2024navigation} 
Thereby, the dynamics models predict various aspects of future states, including positions~\cite{kim2022learning}, visual observations~\cite{bar2024navigation}, and terrain properties such as bumpiness~\cite{kahn2021badgr} or slippage~\cite{gibson2024dynamics, chen2024identifying}, leveraging proprioceptive labels. 
Lately, world models have emerged, which encode system dynamics in a latent space, enabling policy optimization through imagined rollouts~\cite{hafnerdream, hansentd}. 
Such models can also be used to directly estimate the next suitable action~\cite{sridhar2024nomad}.
The prior works dominantly focus on wheeled robots~\cite{xiao2021learning, pokhrel2024cahsor, kahn2021badgr, gibson2024dynamics} or short-horizon predictions~\cite{Duong2024PortHamiltonianNO, kim2023bridging}, often neglecting to incorporate proprioceptive data in the observation space for terrain assessment. Pioneering work targeting quadrupedal robots learned an FDM in simulation with a $2D$ LiDAR scan as observation~\cite{kim2022learning}. While achieving navigation in narrow environments, open challenges remain to go beyond flat scenes with $2D$ obstacles and to investigate the transfer to reality. 
Our approach advances dynamics modeling for quadrupeds by incorporating proprioceptive history and height scans, enabling long-horizon predictions in rough terrain. We improve the sim-to-real transfer performance by integrating real-world data, which in turn results in robust navigation in unstructured terrains.

\subsection{Planning}

Classical planning frameworks employ a modular structure, combining mapping, traversability assessment, and sampling- or optimization-based planning~\cite{beyer2024risk, meng2023terrainnet, frey2023fast, castro2023does, wellhausen2023artplanner, mattamala2022efficient, chavez2018learning, frey2024roadrunner}. Traversability is evaluated either through heuristics~\cite{wellhausen2021rough} or learned from experience~\cite{frey2023fast, meng2023terrainnet, beyer2024risk}. Motion planning techniques such as MPPI~\cite{williams2017information} or iCEM~\cite{PinneriEtAl2020iCEM} sample action sequences, propagate them using dynamics models, and select actions based on traversability and task-specific reward functions. 
While effective, the MPPI formulation requires extensive tuning and environment-specific adjustments~\cite{meng2023terrainnet, frey2024roadrunner, williams2017information}. Our perceptive FDM mitigates this by implicitly learning traversability and directly providing risk scores for action sequences, eliminating the need for manual assessment while retaining the flexibility of sampling-based planning.

End-to-end learning approaches optimize planning policies via unsupervised learning~\cite{roth2024viplanner, yang2023iplanner} or reinforcement learning (RL)~\cite{lee2024learning, hoeller2021learning}, directly mapping sensor inputs to motion commands or paths. These methods offer fast inference and avoid error accumulation across modules. While unsupervised approaches rely on simplified dynamics and require manual cost-map tuning, RL-based planners learn platform-aware behaviors through experience but face sim-to-real transfer challenges due to domain discrepancies.
Our method addresses domain discrepancies by incorporating real-world data into the dynamics model while maintaining platform awareness through learning from past experiences. Additionally, it preserves the benefits of sampling-based planning, allowing flexible adaptation of planning behavior without the need for retraining.

\begin{figure*}[ht!]
    \centering
    \includegraphics[width=1.0\textwidth]{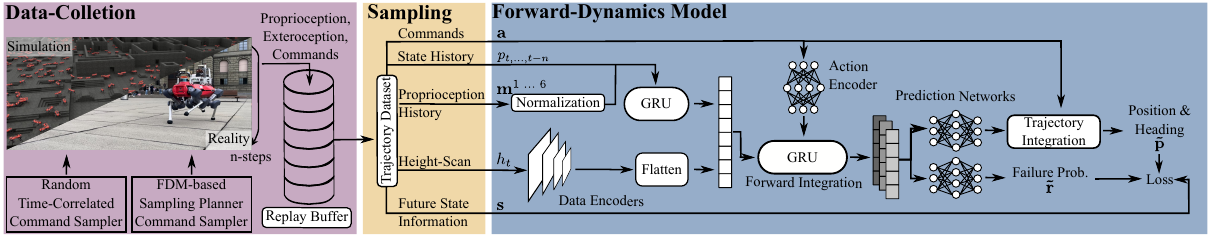}
    \caption{Overview of the FDM training. Data is collected in a parallelized simulation setting and from real-world experiments. The proprioceptive and exteroceptive measurements, along with velocity actions, are saved in a replay buffer from which training data is sampled. The information about the current and past state of the robotic system is encoded and given to a recurrent unit, which generates a latent of the robot's future states conditioned on the applied actions. Different heads are used to predict the future SE2 poses and failure probabilities.}
    \label{fig:overview_training}
\end{figure*}

\begin{figure}[ht!]
    \centering
    \includegraphics[width=1.0\columnwidth]{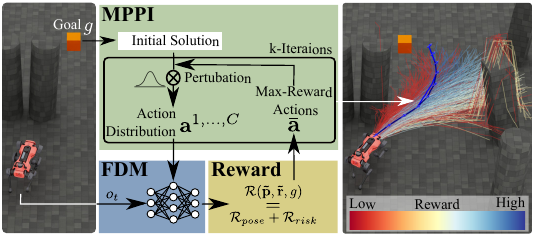}
    \caption{Overview of the MPPI-based planning approach. A population of action trajectories is generated by perturbating an initial solution with Gaussian noise. The presented FDM is then used to predict the future states and the risk of the individual action sequences, which are evaluated using a reward formulation. After $k$ iterations, using the previous highest reward action sequence as a starting point, the sequence with the maximum reward is executed.}
    \label{fig:overview_planning}
\end{figure}

\section{Preliminaries}
\label{sec:preliminaries}

\subsection{Dynamics Modeling} 

We adopt the Partially Observable Markov Decision Process (POMDP) framework to model the system's dynamics. A POMDP is defined as a tuple $(\mathcal{S}, \mathcal{A}, \mathcal{T}, \mathcal{O}, \mathcal{Z})$.
Here, $\mathcal{S}$ represents the set of states, capturing the possible configurations of the system and its environment, while $\mathcal{A}$ denotes the set of actions available to the agent. The state transition probabilities, $\mathcal{T}(s_{t+1} | s_t, a_t)$, describe the likelihood of transitioning from state $s_t$ to $s_{t+1}$ given an action $a_t$. Observations, drawn from the set $\mathcal{O}$, provide partial and noisy state information, with $\mathcal{Z}(o_t | s_t)$ specifying the observation probabilities, which reflect the likelihood of receiving an observation $o_t$ given the state $s_t$.
The robot evolves according to a forward dynamics model that maps the current state and action to the next state:
\begin{equation}
    s_{t+1} = f(s_t, a_t).
\end{equation}

However, such model $f$ is unknown due to the unobservability of the true state $s_t$. 
To address this, we aim to learn an approximate dynamics model $\tilde{f}$ that predicts a subset of state $\tilde{s}$ based on the action $a_t$ and observation $o_t \in \mathcal{O}$:

\begin{equation}
    \tilde{s}_{t+1} \approx \tilde{f}(o_t, a_t).
\end{equation}

Instead of rolling out this model, which would require learning the mapping from $\tilde{s}_t$ to $o_t$, one can extend this one-step model to an $n$-step prediction model for long-horizon forecasting. 
Given the current observation $o_t$ and sequence of actions $\mathbf{a} = a_{t, \dots, t+n-1}$ we predict a sequence of future states $\mathbf{\tilde{s}} = \tilde{s}_{t+1, \dots, t+n}$.
\begin{equation}
    \label{eq:n-step-fdm}
    \mathbf{\tilde{s}} \approx \tilde{f}(o_t, \mathbf{a}).
\end{equation}
In practice, this mitigates the computational complexity associated with long-horizon predictions.

\subsection{Model Predictive Path Integral Control}
This work incorporates the \gls{mppi} control framework to select action sequences $\mathbf{a}$. The selection over a set of $C$ candidates is performed by maximizing a reward function $\mathcal{R}$ defined over the future states $\mathbf{\tilde{s}}$, and the goal pose $g$. Therefore, the action sequences must be forward-propagated through the system's dynamics over a prediction horizon $n$ to compute the future states.

\begin{align}
    \label{eq:bar-a-over-reward}
    \bar{\mathbf{a}} &= \arg\max_{i \in [1, C]} \mathcal{R}(\mathbf{\tilde{s}}^i, g) \\
    &= \arg\max_{i \in [1, C]} \mathcal{R}(\tilde{f}(\mathbf{a}^i, o_t), g)
\end{align}

The optimal action sequence is iteratively refined by perturbing the previous solution with Gaussian noise, evaluating the new set of candidates, and then performing a weighted update:

\begin{equation}
    \label{eq:mppi-action-update}
    \bar{\mathbf{a}} \leftarrow \bar{\mathbf{a}} + \sum_{i=1}^{C} w_i \delta \mathbf{a}_{i},
\end{equation}

where $w_i$ denotes the weight assigned to the $i$-th trajectory and $\delta \mathbf{a}_{i}$ the action pertubation. These weights are computed based on the reward $\mathcal{R}_i$ of each trajectory, ensuring higher-reward trajectories contribute more significantly to the update:

\begin{equation}
    w_i = \frac{\exp \Bigl(\frac{1}{\gamma}(\mathcal{R}_i - \mathcal{R}_{\max})\Bigr)}{\sum_{j=1}^C \exp \Bigl(\frac{1}{\gamma}(\mathcal{R}_j - \mathcal{R}_{\max})\Bigr)},
\end{equation}

where $\gamma$ controls sensitivity to reward differences, and $\mathcal{R}_{\max}$ represents the maximum reward among all sampled trajectories.

\section{Problem Statement}

\subsection{Dynamics Modeling}

To enable accurate state predictions, we approximate the $n$-step transition function introduced in Eq.~\ref{eq:n-step-fdm} with a learned model $\tilde{f}_{\theta}$, parameterized by neural network weights $\theta$. 
We define the state $\tilde{s}$ to be the tuple $(p,r)$, where $p \in \text{SE}2$ is the robot's pose and $r \in \{0,1\}$ is the failure risk of the trajectory where $0$ indicates risk-free and $1$ a catastrophic failure. 
The actions $a \in R^3$ are defined as the linear and angular velocity in the x, y, and yaw direction. 
As observations~$o$ (detailed in Tab.~\ref{tab:observations}), the model utilizes proprioceptive inputs, including the robot’s past states $\tilde{s}_{t, \dots, t-n}$ and measurements $\mathbf{m}^{1, \dots, 7} = m_{t, \dots, t-n}^{1, \dots, 7}$, along with the current height scan $h_t$ as exteroceptive input, enabling perceptive predictions of the robot’s interaction with its environment.
Accordingly, the dynamics model estimates the future poses $\mathbf{\tilde{p}} = \tilde{p}_{t, \dots, t+n}$ and failure risks $\mathbf{\tilde{r}} = \tilde{r}_{t, \dots, t+n}$, taking into account the platform's capabilities, applied locomotion policy, and environmental factors such as friction and terrain roughness. 
Similar to~\cite{lee2023learning}, the neural network is parameterized using a residual formulation. Instead of predicting direct pose estimates, the model predicts residual velocities internally $\Delta\mathbf{a}$ and integrates the final velocity trajectory using a constant-velocity model to final pose estimates.
Consequently, the objective of the dynamics model becomes minimizing a combined loss comprising pose prediction $\mathcal{L}_{pose}$ and failure risk prediction $\mathcal{L}_{risk}$:  
\begin{equation}
    \theta^* = \arg\min_{\theta} \left(\mathcal{L}_{pose} + \mathcal{L}_{risk}\right),
\end{equation}
where $\theta^*$ denotes the optimized model parameters.

\begin{table}[tb!]
    \centering
    \begin{tabular}{llll}
        \hline
        \textbf{\#} & \textbf{Observation} & \textbf{Dimensions} & \textbf{Augmentation} \\
        \hline
        $m^1$ & Twist Commands & $\mathbb{R}^{3}$ & - \\
        $m^2$ & Projected Gravity & $\mathbb{R}^{3}$ & $\mathcal{U}[-0.05, 0.05]$\\
        $m^3$ & Base linear velocities & $\mathbb{R}^{3}$ & $\mathcal{U}[-0.1, 0.1]$\\
        $m^4$ & Base angular velocities & $\mathbb{R}^{3}$ & $\mathcal{U}[-0.2, 0.2]$\\
        $m^5$ & Joint positions & $\mathbb{R}^{b}$ & $\mathcal{U}[-0.01, 0.01]$ \\
        $m^6$ & Joint velocities & $\mathbb{R}^{b}$ & $\mathcal{U}[-1.5, 1.5]$\\
        $m^7$ & Last Two Joint actions & $\mathbb{R}^{2b}$ & - \\
        $h$ & Height Map & $\mathbb{R}^{u \times v}$ & $\mathcal{U}[-0.1, 0.1]$ \\
        \hline
    \end{tabular}
    \caption{The observation space of the FDM combines proprioceptive information of the robot state $m^{2 \cdots 4}$ and the joint states $m^{5 \cdots 7}$ with exteroceptive measurements $h$. $b$ represents the robotic system's joint count, and $u \times v$ is the dimension of the height map. $\mathcal{U}$ indicates uniform distributions used to augment the measurements and make the system robust against sensor noise.}
    \label{tab:observations}
\end{table}

\subsection{Planning}
The planning objective is to identify a safe, collision-free, and efficient sequence of actions $\bar{\mathbf{a}}$ to navigate the robot from its current pose $p_t$ to a goal pose $g$. In this work, the navigation task must be performed online, relying only on onboard sensing and computing. In the proposed method, the optimal action sequence $\bar{\mathbf{a}}$ is determined as defined in Eq.~\ref{eq:bar-a-over-reward} with a reward function compromising position error $\mathcal{R}_{pose}$ and failure risk $\mathcal{R}_{risk}$ given the future states generated by the developed FDM~$\tilde{f}_{\theta}$.

\section{Methodology}
\label{sec:methodology}

We outline the data collection process and sampling strategy in Sec.~\ref{subsec:data_collection_sampling}. The model architecture, designed for computational constraints and noisy observations, is described in Sec.~\ref{subsec:model_architecture}. Training procedures, including the loss formulation for long-horizon accuracy, are detailed in Sec.~\ref{subsec:fdm_loss}. An overview of the entire FDM training procedure is provided in Fig.~\ref{fig:overview_training}. The planner's workflow, leveraging MPPI control, is depicted in Fig.~\ref{fig:overview_planning}. The reward formulation, including goal reward and risk penalization, is presented in Sec.~\ref{subsec:planning_loss}.

\subsection{Data Collection and Sampling}
\label{subsec:data_collection_sampling}

\paragraph*{Data Sources} The training data of the Forward Dynamics Model - consisting of the proprioceptive and exteroceptive observations and the states - can be collected from trajectories executed in both simulated environments and during real-world deployments. As the data is conditioned on the specific platform and applied locomotion policy, the specific terrain interactions are captured.
Collecting data in simulation allows for cheap, scalable data generation from thousands of robots in parallel with terrain randomization to achieve wider generalization and robustness. While the simulation provides only a simplified dynamics model, it enables data generation of risky maneuvers that would severely damage the real platform. On the contrary, real-world data is more expensive to collect and only covers safe paths. However, as we later show in Exp.~\ref{subsec:exp-real-world-fine-tune}, this data source remains essential in order to remove undesirable biases, cover the actual sensor noise, and include the full dynamics of the platform, especially in environments beyond the rigid-body domain of current simulators (e.g., snow or entanglement in soft vegetation). Thus, such data enables a closer FDM to the real hardware.

\paragraph*{Synethic Data Generation} To minimize the gap between simulated and real data, we identify the simulation parameters using system identification or model certain parts using learned networks fitted from real-world data~\cite{hwangbo2019learning}. Similarly, the locomotion policies executed in the simulation are the same as those later used in the real system. Moreover, the randomization of observations (see Tab.~\ref{tab:observations}) and terrains increases the diversity of covered state transitions. The action generation is adapted over time. During the initial phase of the model training, actions are purely generated using linear and normal time-correlated sequences, as introduced in~\cite{kim2022learning} and detailed in Appendix~\ref{subsec:time-correlated-action-sampling}.
In later stages, the MPPI planner using the currently trained FDM generates a part of the action sequences given randomly sampled goal poses. This allows the model to adjust for the different sampling nature of the planner and avoid possible overfitting to the time correlation. 

\paragraph*{Real-World Data Collection} 
A human expert operator safely guides the ANYmal robot equipped with the Boxi \cite{Frey-Tuna-Fu-RSS-25} sensor payload through a variety of environments. During this data collection, proprioceptive inputs and height-scan generated from onboard depth cameras are stored. The accurate pose estimation of the robot is provided by fusing dual RTK-GNSS, highly accurate IMU measurements, and position estimates provided by a Leica Geosystems MS60 Total Station using the open-source Holistic Fusion factor graph framework \cite{nubert2025holistic}.

\paragraph*{Data Sampling}
The states and the observations are stored in replay buffers from which arbitrary sample numbers can be generated. While the buffers remain constant for real-world deployments, buffers are cleared and newly populated per episode of the simulation. 
As the buffers contain trajectories multiple times longer than the prediction horizon, we decompose them into sub-trajectories, starting at random timestamps in the trajectory. To create training samples, for each sub-trajectory, the history information as part of the observations is collected for $n$ past states with a frequency of $1 / \Delta t_h$.
This history horizon of $n \cdot \Delta t_h$ enables the model to infer terrain properties such as roughness by observing the platform's recent motion and interaction history.
The future states are collected from the following $n$ states with a frequency of $1/ \Delta t_p$, resulting in a prediction horizon of $n \cdot \Delta t_p$.
For all samples, the poses of the observation history and future states are translated into the robot's base frame at time $t_0$.

\begin{figure*}
    \begin{tikzonimage}[width=\textwidth]{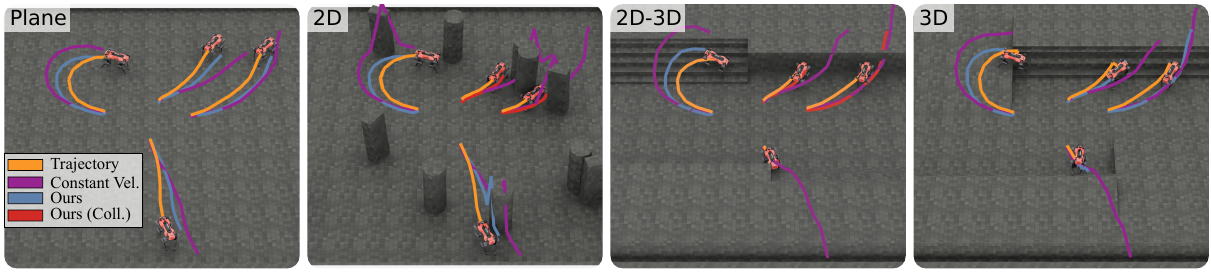}[image label_bl]
        \node{(a) Sim};
    \end{tikzonimage}
    \begin{tikzonimage}[width=\textwidth]{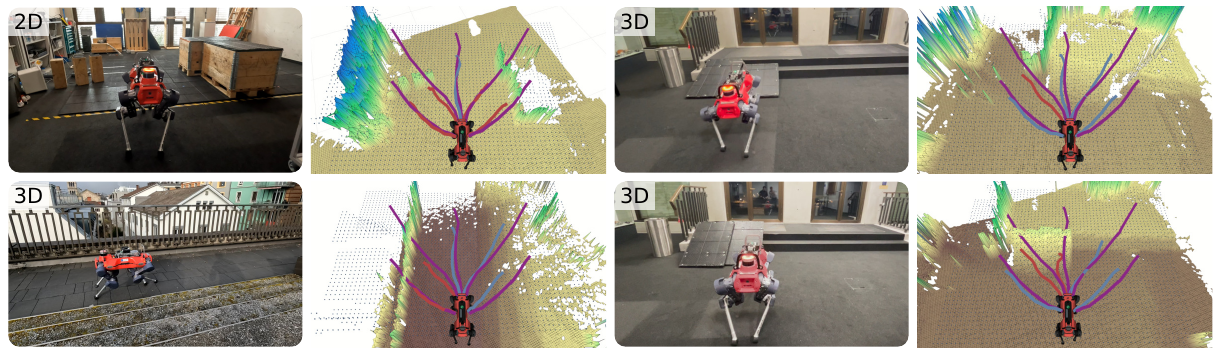}[image label_bl]
        \node{(b) Real};
    \end{tikzonimage}
    \caption{Demonstration of environment- and platform-aware state predictions using the presented FDM. Collision-free predictions of our method are displayed in \textcolor{ours}{\rule{1.5ex}{1.5ex}}, in collision ones in \textcolor{collision}{\rule{1.5ex}{1.5ex}}, whereas the actual path is presented in  \textcolor{trajectory}{\rule{1.5ex}{1.5ex}}. (a) Simulation: The same four action sequences are rolled out across multiple environments, showing that the predicted paths by our model are adapted to the environment. (b) Real-World: Qualitative comparison between constant velocity estimation \textcolor{constant_vel}{\rule{1.5ex}{1.5ex}} and our model's predictions for the same action sequences across multiple scenarios.
    Given that rigid body dynamics sufficiently describe the scene, we deployed our synthetic model to assess sim-to-real transfer.}
    \label{fig:test_perceptive_fdm}
\end{figure*}

\subsection{Model Architecture}
\label{subsec:model_architecture}

\paragraph*{Model Input} 
As introduced, the model receives as observations a history of $n$ past states $\tilde{s}_{t, \dots, t-n}$,  proprioceptive readings $m^{1, \dots, 7}_{t, \dots, t-n}$ and a height scan $h_{t}$ for traversability assessment and obstacle detection.
From the proprioceptive data of all samples, mean and standard deviation are computed to normalize these observations before feeding them to the network.
The noise augmentation, as detailed in Tab.~\ref{tab:observations}, for the synthetic samples of the proprioceptive and exteroceptive observations follows~\citet{rudin2022learning}. 
In addition, the height scan has been augmented with missing patches, and all its occlusions have been specifically modeled to capture the limitations of the real-world measurement. Occlusions are determined by checking for a direct line of sight between any of the robot depth cameras and the height scan point.

\paragraph*{Model Structure}
An initial GRU layer sequentially encodes the history information of the past states and proprioceptive measurements, while multiple convolutional layers process the current height scan. 
The flattened output of the latter and the last embedding of the former are used to initialize the hidden state of the forward prediction GRU. 
This unit receives the sequence of action encodings from an MLP and sequentially predicts a latent for each future state.
All future state encodings are processed in parallel by two prediction heads. 
There is one head to predict the twist command corrections $\Delta\mathbf{\tilde{a}}$ and another to estimate the failure risk $\mathbf{\tilde{r}}$.
We use two different GRU units, given that the history and prediction frequency differ.
The correction term $\Delta\mathbf{\tilde{a}}$ describes the difference between the intended velocity and the applied one on the robot $\mathbf{\hat{a}}$, s.t. $\mathbf{a} + \Delta\mathbf{\tilde{a}} \approx \mathbf{\hat{a}}$.
Final poses $\mathbf{\tilde{p}} \in \mathbb{R}^2$ are derived by integrating applied velocities over the prediction timestep $\Delta t_{p}$.

\subsection{FDM Loss}
\label{subsec:fdm_loss}

The Forward Dynamics Model loss $\mathcal{L}$ consists of supervised terms for network outputs. Labels are generated from the replay buffer structure introduced in Sec.~\ref{subsec:data_collection_sampling}, using the future states of the trajectories.
The pose loss is computed using mean squared error (MSE) between predicted and true poses. For the heading, a sinus-cosine encoding is applied to avoid discontinuities at $2\pi$. The ground truth poses for failure trajectories is kept constant from the moment the failure occurred. The failure risk is supervised using binary cross-entropy loss (BCELoss) over the trajectory. Therefore, the loss terms are defined as:

\begin{align}
    \mathcal{L}_{pose} (p, \tilde{p}) &= \text{MSELoss}(p, \tilde{p}) \\
    \mathcal{L}_{risk} (r, \tilde{r}) &= \text{BCELoss}(r, \tilde{r})
\end{align}

Additionally, when a failure is predicted, the pose should remain constant for the future trajectory. To highlight this behavior, an extra stop loss $\mathcal{L}_{stop}$, implemented as MSELoss, is applied for failure scenarios:

\begin{equation}
    \mathcal{L}_{stop} = \text{MSELoss}(\tilde{p}, p) \quad \forall p_t \quad \text{with} \quad r_t > \delta_{risk},
\end{equation}

with $\delta_{risk}$ as the threshold to declare the future trajectory as risky. The final loss for model updates becomes a weighted sum of all individual terms:

{\small
\begin{equation}
        \mathcal{L} = \epsilon_{pose} \cdot \mathcal{L}_{pose} + \epsilon_{risk} \cdot \mathcal{L}_{risk} + \epsilon_{stop} \cdot \mathcal{L}_{stop},
\end{equation}
}

where $\epsilon_{pose}$, $\epsilon_{risk}$, and $\epsilon_{stop}$ are the individual weights. Details on how the training is structured and about the iterative approach between data collection and model updates are provided in Sec.~\ref{par:model-training}.

\subsection{Path Planning}
\label{subsec:planning_loss}

Using a zero-shot MPPI planner allows for adjustments of the planning behavior without retraining. Leveraging the pose and failure risk of the perceptive FDM, there are no requirements for handcrafted cost-maps or other metrics to account for the traversability of the environments. Consequently, the planning reward $\mathcal{R}$ can be simplified to a weighted combination of a goal-oriented pose reward $\mathcal{R}_{pose}$ (terminal reward) and a risk minimization term $\mathcal{R}_{risk}$ (state reward). Both components are assigned a weight, $\lambda_{pose}$ and $\lambda_{risk}$, to balance their influence:

\begin{equation}
    \mathcal{R} = \lambda_{pose} \cdot \mathcal{R}_{pose} + \lambda_{risk} \cdot \mathcal{R}_{risk}
\end{equation}

The pose reward $\mathcal{R}_{pose}$ encourages paths that reduce the Euclidean distance between the predicted terminal pose of the robot $p_{t+n}$, and the goal pose $g$. To create a pull factor when being close to the goal, a reward multiplier $\lambda_{pull}$ is applied when the state is closer than a threshold $\delta_{pose}$.

{\scriptsize
\begin{equation}
        \mathcal{R}_{pose} (p_{t+n}, g) = 
        \begin{cases} 
            \| p_{t+n} -  g \|_2 \cdot \lambda_{pull}, & \text{if } \| p_{t+n} -  g \|_2 < \delta_{pose} \\ 
            \| p_{t+n} -  g \|_2, & \text{else}.
        \end{cases}
\end{equation}
}

The risk term $\mathcal{R}_{risk}$ penalizes trajectories with high predicted failure risk $\mathbf{r}_t$, which the FDM estimates based on terrain and motion characteristics. If a path's risk exceeds the threshold $\delta_{risk}$, a penalty $\lambda_{risk}$ is applied. To enhance robustness against false negatives in the risk prediction, the total cost includes the cumulative risk of $q$ neighboring paths. This redundancy increases robustness to isolated collision prediction errors.

\begin{equation}
    \mathcal{R}_{risk} (\mathbf{r}) = \sum_i^q \mathbf{r}^{i} \cdot 
    \begin{cases}
        \lambda_{risk} & \text{if} \quad \exists r_t \in \mathbf{r}^i, \; r_t > \delta_{risk}\\
        0 & \text{else }
    \end{cases}
\end{equation}

\begin{figure*}[ht!]
    \centering
    \includegraphics[width=1.0\textwidth]{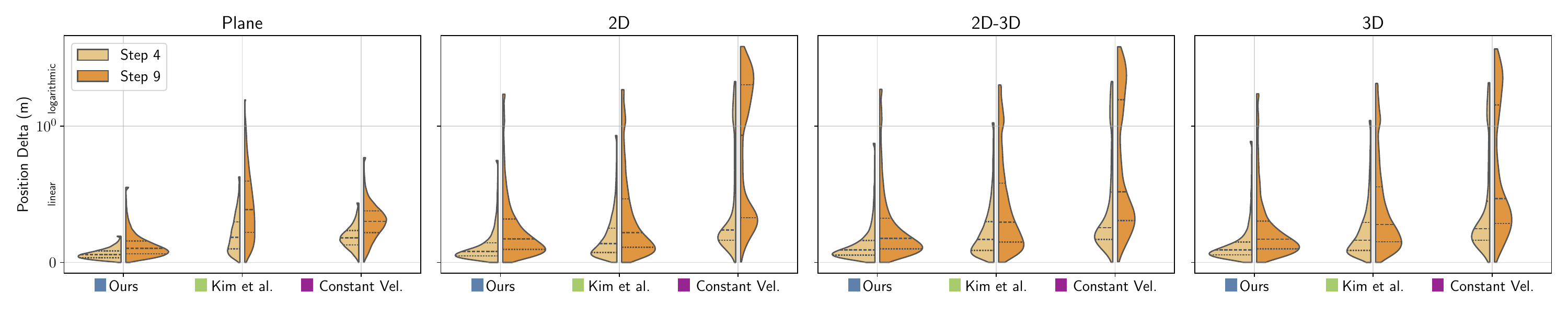}
    \caption{Comparison of the position error at the final prediction step in different environments for the presented FDM  \textcolor{ours}{\rule{1.5ex}{1.5ex}}, the perceptive FDM by~\citet{kim2022learning}  \textcolor{baseline}{\rule{1.5ex}{1.5ex}} and the constant velocity model  \textcolor{constant_vel}{\rule{1.5ex}{1.5ex}}. For each environment, 50k samples are evaluated, and the error is displayed up to the 95\% quantile to mitigate the effect of outliers. The presented method achieves the lowest error rates in all environments. Notably, while the perceptive baseline exhibits an error increase in $3D$ environments, the developed FDM is unaffected by the more complex obstacles.}
    \label{fig:performance-final-step}
\end{figure*}

\section{Experiments}
\label{sec:experiments}

\paragraph*{Experimental Setup}
The effectiveness and perceptive capabilities of the developed FDM are evaluated in both simulated and real-world environments. In simulation, experiments are conducted in three scenarios: $2D$, $2D$-$3D$, and $3D$. $2D$ environments include obstacles like walls, pillars, and mazes, detectable with $2D$ sensors, while $3D$ scenarios feature complex obstacles such as stairs and ramps. These obstacles cannot be differentiated from walls using only a horizontal $2D$ sensor without actively changing the observation angle. Consequently, at least $2.5D$ representations would be required. $2D$-$3D$ environments combine both obstacle types. Large-scale simulations are performed on the legged robot ANYmal~\cite{hutter2016anymal}, Barry~\cite{valsecchi2023barry}, and the wheeled-legged robot ANYmal-On-Wheels (AoW)~\cite{bjelonic2019keep}. The simulation results are achieved by building upon the NVIDIA IsaacLab framework~\cite{mittal2023orbit} with terrain details and data augmentations provided in Appendix~\ref{subsec:terrain_details}. Real-world data is collected using ANYmal, which is also used in our real-world deployments.
The FDM runs onboard using an NVIDIA Jetson Orin AGX, with the planner running at $7\,\mathrm{Hz}$ using 2048 trajectories and a model inference time of \SI{40.6}{ms} per iteration. Throughout the experiment section, we use consistent color coding of our method \textcolor{ours}{\rule{1.5ex}{1.5ex}}, the baseline of~\citet{kim2022learning} \textcolor{baseline}{\rule{1.5ex}{1.5ex}} and the constant velocity assumption \textcolor{constant_vel}{\rule{1.5ex}{1.5ex}}.

\paragraph*{Model Training}
\label{par:model-training}
The model is trained to predict the following ten states with a step time of $\Delta t_p = 0.5\,\mathrm{sec}$, yielding a $5\,\mathrm{sec}$ prediction horizon. The history information of the past ten states is collected with a step time of $\Delta t_h = 0.05\,\mathrm{sec}$. Training alternates between data generation and model updates to ensure diverse coverage. Initially, only synthetic data is used to create a robust model through broad environmental variability and data augmentation. Across 15 rounds, each collecting 80k samples from 10k parallel environments, updates consist of 8 episodes with a batch size of 2048, optimized using the AdamW optimizer with a learning rate scheduler and weight decay. In later stages, real-world data is integrated with synthetic data, and weights are refined using a small, constant learning rate to capture the full system dynamics beyond the rigid-body domain. Training is performed on a single NVIDIA RTX 4090, completed in approximately eight hours. Subsections \ref{subsec:exp-fdm-percetive} to \ref{subsec:exp-platform-aware}, \ref{subsec:exp-planning-perf}, and \ref{subsec:exp-planning-real-world} utilize models trained exclusively on synthetic data, while subsection \ref{subsec:exp-real-world-fine-tune} employs fine-tuned models incorporating real-world data. More details on the sensitivity of learning and planning parameters, alongside a discussion of the adaptation required for a new robot platform, can be found in Appendix \ref{app:adaptation}.

\subsection{FDM Perceptiveness}
\label{subsec:exp-fdm-percetive}
By incorporating both proprioceptive and exteroceptive measurements, the proposed method can predict robot-terrain interactions. Specifically, the FDM can estimate failure states (e.g., collisions) and adjust future poses based on the velocity command tracking performance in rough terrain.
To evaluate its perceptiveness, we apply the same action sequence across different terrains and visualize the resulting paths in Fig.~\ref{fig:test_perceptive_fdm}. Even on flat ground, the simple constant velocity assumption fails to capture the robot’s actual dynamics, whereas our approach closely aligns with the walked path. The advantage of our method becomes even more apparent in complex environments, where it accurately detects collisions and models movement on stairs and ramps. A similar performance is observed in real-world scenarios, demonstrating the model’s sim-to-real transfer capabilities.

\begin{figure}[t!]
    \centering
    \includegraphics[width=0.8\columnwidth]{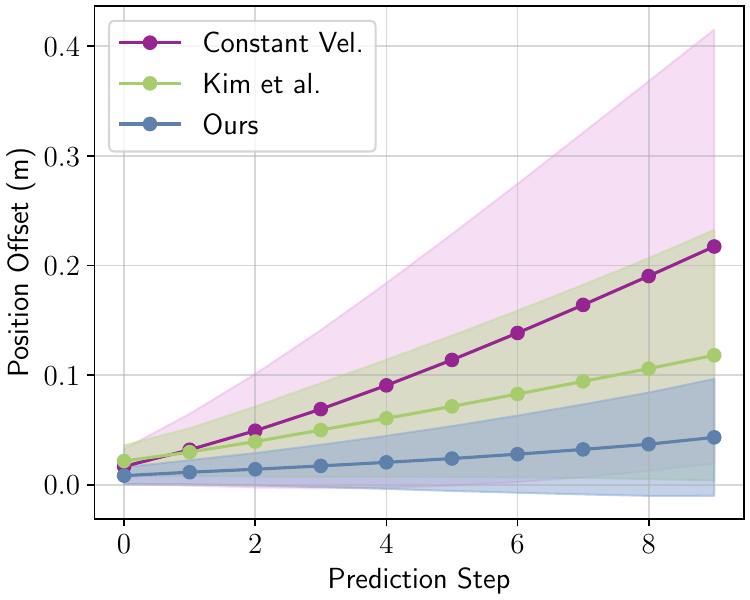}
    \caption{Comparison of the position error over the prediction steps between the presented method, the perceptive FDM by~\citet{kim2022learning}, and the constant velocity model. Our FDM demonstrates the highest accuracy with the lowest errors and smallest standard deviation.}
    \label{fig:performance-over-time}
\end{figure}

\begin{figure*}[ht!]
    \includegraphics[width=1.0\textwidth]{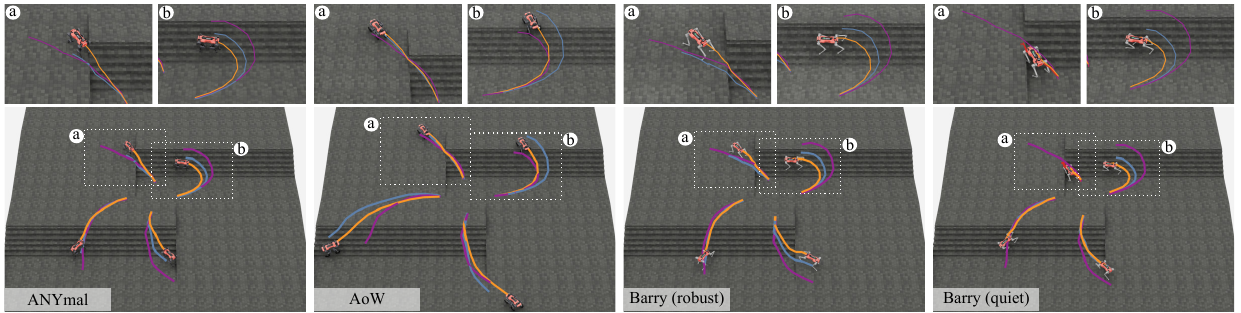}
    \caption{Comparison of state predictions of the presented method on the quadrupedal platforms ANYmal~\cite{hutter2016anymal}, Barry~\cite{valsecchi2023barry}, and ANYmal-On-Wheels (AoW)~\cite{bjelonic2019keep}. For Barry, a robust locomotion policy capable of traversing rough environments and a "quiet" locomotion policy optimized for minimal torques in mostly flat environments have been deployed. Moreover, the height-scan size has been extended for AoW to account for the wider movement range. While all platforms receive the same actions, as visible in the constant velocity  \textcolor{constant_vel}{\rule{1.5ex}{1.5ex}} predictions, the platforms display different dynamics due to changing structure and actuators (see their trajectory \textcolor{trajectory}{\rule{1.5ex}{1.5ex}}). The presented FDM \textcolor{ours}{\rule{1.5ex}{1.5ex}} demonstrates platform-aware predictions, successfully capturing the platform changes.}
    \label{fig:platform-aware}
\end{figure*}

\subsection{Baseline Comparison}

We evaluate our proposed method quantitatively by running experiments on a larger scale against the baseline method of~\citet{kim2022learning} and the constant velocity assumption. The baseline relies on a 2D-Lidar scan as exteroceptive information and predicts future positions and collisions. While we keep the original model structure, loss formulation, and training hyperparameters, we train the baseline with the same data as our model to ensure comparability.
The evaluation includes 50k samples collected from each of the previously introduced environments.
In Fig.~\ref{fig:performance-over-time}, we demonstrate that the position error averaged over all environments remains the smallest for the developed FDM with a decrease of 41.28\% compared to the perceptive baseline and 70.57\% compared to the constant velocity assumption in the final prediction step. Regarding the position error in the individual environments, displayed in Fig.~\ref{fig:performance-final-step}, it is evident that most predictions of our method are within small error regions. 
Nevertheless, large error outliers exist where, e.g., the failure prediction is wrong, leading either to overshooting or discontinuing paths.
The training terrain itself contains flat patches, however, the perceptive baseline exhibits issues to generalize in this case. In contrast, our model provides accurate predictions in the tested scenario. Further, the better accuracy compared to the baselines becomes clearly evident when moving towards the $3D$ environments, where the restricted $2D$ LiDAR scan does not provide sufficient information. 
Regarding the collision estimation, the developed FDM demonstrates an accuracy of at least 89\% over all environments. Our method predicts collision in environments with $2D$ obstacles correctly with an F1 score of 0.9, with only a minor decrease to 0.85 when applied in $3D$ environments (see Tab.~\ref{table:methods_env}). 
The baseline achieves a higher recall score, which we hypothesize is due to its limited perception. This restriction prevents effective differentiation of obstacles, such as stairs and walls, leading to more conservative failure predictions and a bias toward false positives. As a result, precision is significantly reduced, ultimately lowering the overall F1 score.
In the planar environment, robot failures are rare (0.042\%), likely caused by simulation instabilities that the model cannot predict, leading to low recall and precision scores.
Overall, the presented method achieves the highest position prediction accuracy over all environments and prediction steps. Moreover, it demonstrates the most precise failure estimation, although it is less likely to detect all collisions compared to the more conservative baseline. 

\begin{table}[t!]
    \centering
    \resizebox{\columnwidth}{!}{%
        \begin{tabular}{>{\centering\arraybackslash}p{0.3cm}lccccc}
            \hline
            \textbf{Env.} & \textbf{Method} & \textbf{Pos. Offset} & \textbf{Precision} & \textbf{Recall} & \textbf{Accuracy} & \textbf{F1 Score} \\
            \hline

            \multirow{3}{*}{\rotatebox{90}{Plane}}
            & Constant Vel. & 0.32 $\pm$ 0.24 & - & - & - & - \\
            & Kim et al. & 0.45 $\pm$ 0.33 & \textcolor{gray}{10.84} & \textcolor{gray}{90.47} & \textcolor{gray}{92.92} & \textcolor{gray}{0.17} \\
            & Ours & \textbf{0.13 $\pm$ 0.19} & \textcolor{gray}{12.73} & \textcolor{gray}{9.83} & \textcolor{gray}{98.32} & \textcolor{gray}{0.10} \\
            \hline
            \multirow{3}{*}{\rotatebox{90}{2D}}
            & Constant Vel. & 1.33 $\pm$ 1.17 & - & - & - & - \\
            & Kim et al. & 0.37 $\pm$ 0.41 & 80.63 & \textbf{92.68} & 86.47 & 0.86 \\
            & Ours & \textbf{0.28 $\pm$ 0.34} & \textbf{93.42} & 86.57 & \textbf{89.13} & \textbf{0.90} \\
            \hline
            \multirow{3}{*}{\rotatebox{90}{2D-3D}}
            & Constant Vel. & 1.08 $\pm$ 1.11 & - & - & - & - \\
            & Kim et al. & 0.45 $\pm$ 0.45 & 70.11 & \textbf{88.97} & 83.61 & 0.78 \\
            & Ours & \textbf{0.30 $\pm$ 0.37} & \textbf{83.09} & 87.25 & \textbf{89.20} & \textbf{0.85} \\
            \hline
            \multirow{3}{*}{\rotatebox{90}{3D}}
            & Constant Vel. & 0.99 $\pm$ 1.06 & - & - & - & - \\
            & Kim et al. & 0.44 $\pm$ 0.45 & 74.05 & \textbf{86.62} & 86.35 & 0.80 \\
            & Ours & \textbf{0.28 $\pm$ 0.35} & \textbf{83.58} & 86.51 & \textbf{90.61} & \textbf{0.85} \\
            \hline
        \end{tabular}
    }
    \caption{Comparison between the developed FDM, a perceptive FDM using a $2D$ LiDAR by~\citet{kim2022learning} and the constant velocity baseline over multiple environments. The presented method demonstrates the lowest final position error and highest failure prediction accuracy over all test environments. For the failure estimation, a \textit{positive} case is a high-risk action sequence, whereas \textit{negative} indicates a safe one. The higher precision scores of our method underline that if it predicts a collision, it is the most likely of all methods to be correct. The perceptive baseline is more conservative and achieves higher recall scores. 
    Failures in the planar environment occur at a rate of 0.042\% (compared to 50–60\% in other cases), likely due to simulation instabilities that the models cannot predict.}
    \label{table:methods_env}
\end{table}

\subsection{Platform-aware Predictions}
\label{subsec:exp-platform-aware}
As the data to train the presented method is sampled from trajectories, the learned models are aware of the capabilities of the platform and its locomotion policy. We train individual FDMs for the above-introduced platforms. Two different locomotion policies have been deployed for Barry: firstly, a robust policy capable of overcoming rough terrain, and secondly, a "quiet" policy used to minimize torques, which is not suitable for rough terrains.  As demonstrated in Fig.~\ref{fig:platform-aware}, given the same action sequence, the FDMs show robot embodiment-specific predictions, capturing the actual robot locomotion capabilities and dynamics. This underlines the model's platform-aware training.

\begin{figure*}[ht!]
    \centering
    \includegraphics[width=1.0\textwidth]{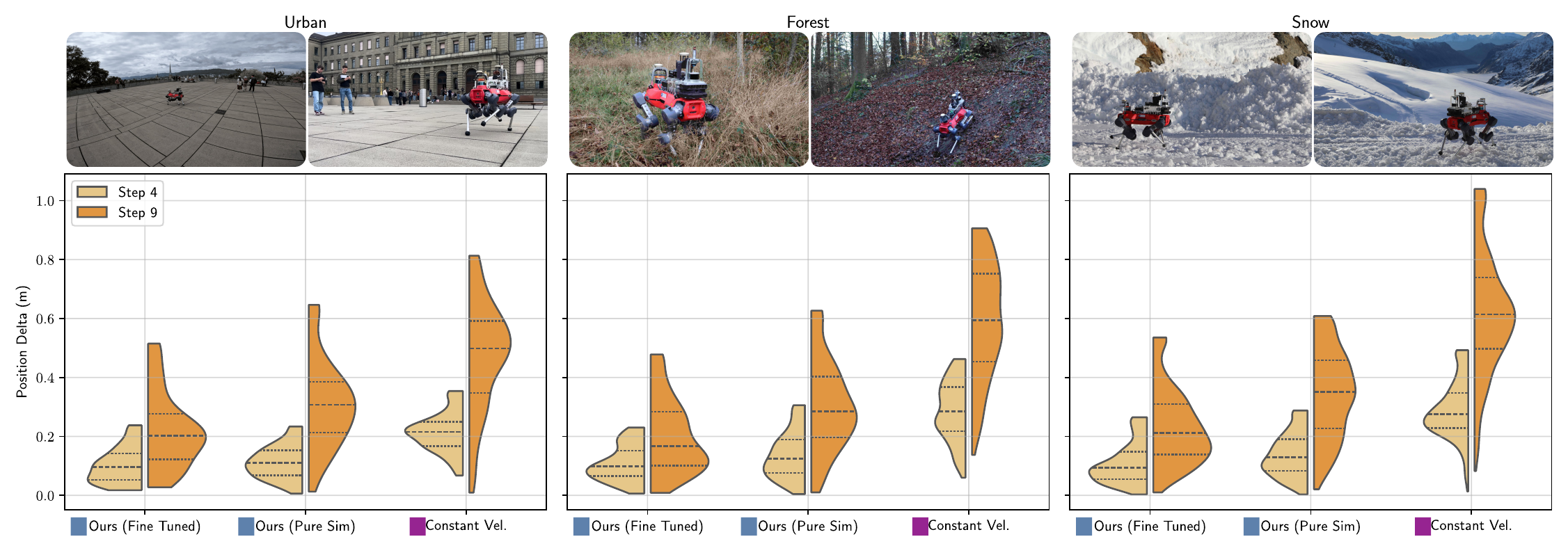}
    \caption{Comparison of the position error at two prediction steps in real-world environments. Shown is the presented method \textcolor{ours}{\rule{1.5ex}{1.5ex}}, trained only with simulated data and fine-tuned with real-world data and the constant velocity model \textcolor{constant_vel}{\rule{1.5ex}{1.5ex}}. The presented method can already bridge successfully to the real world. However, it still exhibits larger errors that our deployed fine-tuning can reduce. 
    }
    \label{fig:real-world-fine-tuning}
\end{figure*}

\subsection{Real-World Fine Tuning}
\label{subsec:exp-real-world-fine-tune}

During the synthetic data generation, we randomize the terrain to increase robustness and simplify the reality transfer. For rigid environments, this allows us to directly transfer our model, as demonstrated in Fig.~\ref{fig:test_perceptive_fdm}.
When targeting dynamic scenarios beyond the rigid domain, the synthetic data becomes out of domain, potentially increasing modeling errors.
As detailed above, we fuse data collected from difficult-to-model real-world deployments into the training to cover soft, slipping, or entangled scenarios. This real-world data mix includes samples from pavement, snow, and forest deployments collected as part of the \textit{GrandTour}~\cite{Frey-Tuna-Fu-RSS-25}.  The snow environment demonstrates frequent slipping events, while the forest environments present challenges such as slipping and entanglement in the high grass and other vegetation.
For the evaluation presented in Fig.~\ref{fig:real-world-fine-tuning}, new datasets in similar environments have been used. The experiments show that even before the fine-tuning, our FDM performs better than the constant velocity model. After fine-tuning, accuracy improves further, reducing the mean position error by 34.38\% in the forest, 30.55\% in the snow mountain scenario, and 30.30\% on the pavement. The similar error reduction on pavement, despite it being less out-of-distribution, can be attributed to the abrupt movements observed, in contrast to the smoother patterns in the other datasets. A comparison to \citet{kim2022learning} is not possible due to the absence of a 2D Lidar in our available datasets. 

\begin{figure*}[ht!]
    \centering
    \includegraphics[width=1.0\textwidth]{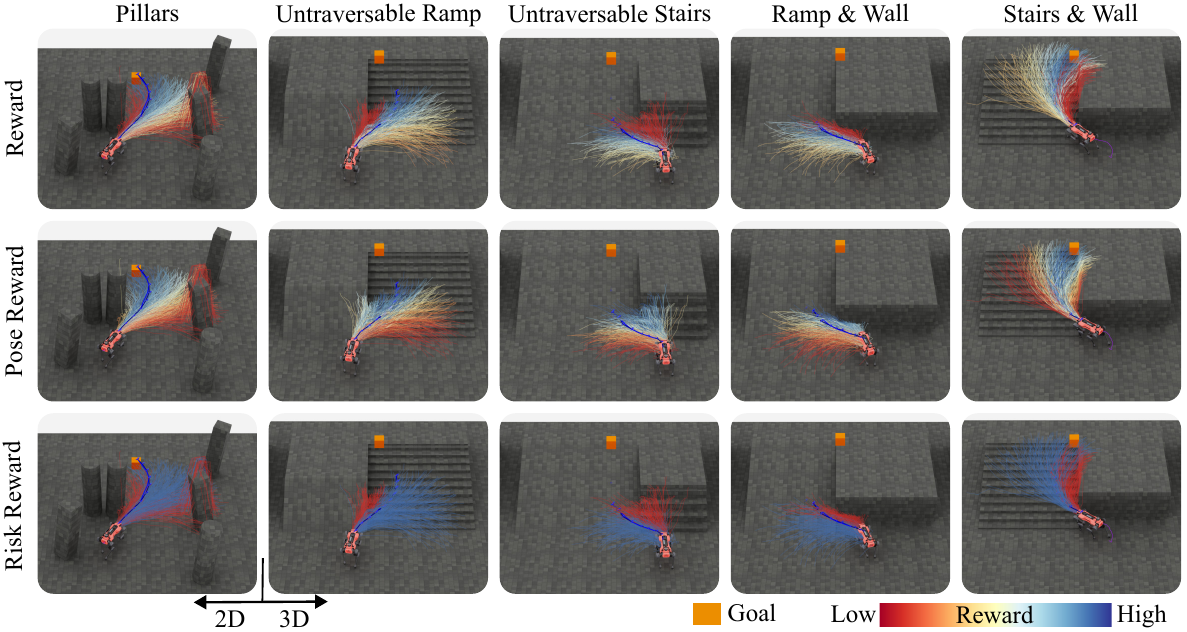}
    \caption{Demonstration of the pose and failure rewards across various simulation scenarios. The proposed FDM accurately predicts failures due to collisions and early path terminations caused by untraversable stairs and ramps. As a result, the simple combination of a pose reward guiding the robot toward the goal and a failure reward preventing collisions proves sufficient for safe and effective planning.}
    \label{fig:planning_sim}
\end{figure*}

\subsection{Planning Performance}
\label{subsec:exp-planning-perf}
To evaluate planning performance, we compare the MPPI planner using the proposed FDM and reward formulation from Sec.~\ref{subsec:planning_loss} to an MPPI planner using the learned FDM method of~\citet{kim2022learning} and the height-scan-based traversability estimation of~\citet{wellhausen2021rough}. In the latter, the failure loss term is replaced by an evaluation of future robot positions on the generated traversability map.
For this experiment, the baseline method of~\citet{kim2022learning} was trained solely in a $2D$ environment, as training in a more complex $3D$ environment made the collision predictions, as expected, unreliable, rendering the FDM unsuitable for path planning. The MPPI parameters have been tuned for each baseline, with details provided in Appendix~\ref{app:expanded_exp}.
We assess the planner's effectiveness in both $2D$ and $3D$ environments based on success rate, mean path length (MPL), and mean path time (MPT). As shown in Tab.~\ref{tab:planning_performance}, our approach achieves the highest success rate across both environments. While the baseline methods perform well in $2D$, their performance drops significantly in $3D$ due to their limited ability to generalize across different obstacle shapes and slopes, such as stairs and ramps, often misclassifying traversable paths as impassable. MPT and MPL scores are reported for all paths and successful paths only. Reporting on all paths enables a comparison across methods using the same dataset, while focusing on successful paths reduces the influence of early-collision trajectories, which tend to lower both metrics. The results show that the presented method produces the most effective paths towards the goal. In contrast to the baseline method by~\citet{kim2022learning}, the average time and length for all paths is typically lower than for successful paths, indicating that our method failed in cases where the goal is not reached. The more conservative baseline instead circled around the obstacles, leading to increased path time and length, often without reaching the goal. 

\begin{table}[t!]
    \centering
    \resizebox{\columnwidth}{!}{ %
        \begin{tabular}{clccccc}
            \hline
            \textbf{Env.} & \textbf{Method} & \textbf{Success (\%)} 
            & \multicolumn{2}{c}{\textbf{MPL (m)}} 
            & \multicolumn{2}{c}{\textbf{MPT (s)}} \\
            & & & \textbf{Suc.} & \textbf{All} & \textbf{Suc.} & \textbf{All} \\
            \hline
            \multirow{3}{*}{2D} 
            & MPPI Ours & \textbf{88.33} & \textbf{4.28} & \textbf{4.02} & \textbf{9.23} & \textbf{8.92} \\
            & MPPI Kim et al.~\citep{kim2022learning} & 78.33 & 9.98 & 16.23 & 25.01 & 41.88 \\
            & MPPI Heuristics~\citep{wellhausen2021rough} & 82.50 & 4.73 & 4.79 & 11.12 & 12.68 \\
            \hline
            \multirow{3}{*}{3D} 
            & MPPI Ours & \textbf{73.75} & \textbf{3.93} & \textbf{3.69} & \textbf{8.68} & \textbf{9.67} \\
            & MPPI Kim et al.~\citep{kim2022learning} & 48.75 & 7.20 & 13.26 & 18.60 & 35.45 \\
            & MPPI Heuristics~\citep{wellhausen2021rough} & 33.13 & 4.41 & 3.85 & 13.99 & 17.17 \\
            \hline
        \end{tabular}
    }
    \caption{Comparison of planning methods in $2D$ and $3D$ environments, evaluating Success Rate, Mean Path Length (MPL), and Mean Path Time (MPT). 
    The MPL and MPT metrics are reported for successful paths reaching the goal and all paths.
    Our approach demonstrates superior performance compared to the baseline method of~\citet{kim2022learning}, which struggles to assess traversability. While the heuristics-based method performs well in the $2D$ case, it fails to generalize to varying height differences of obstacles in $3D$ and would require fine-tuning for each obstacle type.
    Further, our approach executes successful paths in the shortest time and path length due to more precise knowledge of the dynamics.}
    \label{tab:planning_performance}
\end{table}

\subsection{Qualitative Planning Evaluation}
\label{subsec:exp-planning-real-world}

We test the planner in simulated and real-world settings to assess the system’s ability to generate safe and efficient trajectories while handling environmental uncertainty, sensor noise, and terrain variability.
As demonstrated in Fig.~\ref{fig:planning_sim}, in simulation, the sampled action trajectories avoid obstacles and untraversable regions while steering toward the goal as our FDM successfully adjusts the future poses based on the terrain and estimates the risk correctly. 
In the real-world deployment, as shown in Fig.~\ref{fig:real_world_planning}, we illustrate a long-range traversed path, where the goal positions are projected onto the robot's perception range at each time step. Despite real-world challenges such as sensor noise, terrain inconsistencies, and imperfect state estimation, our FDM successfully interprets the environment's traversability. This demonstrates that using the proposed FDM, safe planning can be achieved while relying only on the two simple cost terms.

\section{Limitations}  
\label{sec:limitations}  

While the presented method demonstrates significant advancements in perceptive dynamics modeling, it is subject to certain limitations. First, despite using data augmentation techniques and including real-world data during training, the model remains constrained to a primarily geometric domain. 
We assume that our method generalizes to terrains with geometric variations covered during training and with terrain properties where the locomotion policy reasonably tracks the velocity command, as demonstrated by our successful real-world experiments. 
However, the FDM is inherently limited by the locomotion policy’s capabilities and may fail in scenarios involving novel geometries, such as spiral staircases, highly confined spaces like caves or tunnels, or extreme terrain conditions like ice or deep mud.
This limitation prevents it from fully capturing the broader dynamics and complexities of diverse real-world scenarios. 
Second, the failure states observed in simulation environments do not perfectly translate to real-world failures, and real-world data lacks demonstrations of collisions due to the risk of hardware damage, leaving a persistent gap between simulation and reality that may affect performance. 
Third, although we significantly reduce the effort required to adjust safety-related parameters in the MPPI-based planning, some tuning is still needed for the action distribution, including command ranges and time correlation factors. In addition, the simulation and learning setup introduces a set of new design choices and tunable parameters. However, in our experiments, the learning setup was robust across a range of hyper-parameters, robot platforms, and simulation environments, while the correct environment design was key for successful sim-to-real transfer. 
Generally, we do not expect the model to generalize to environments beyond the training domain. Designing diverse environments that support broader generalization across unseen scenarios remains an open research question.
Finally, our method does not consider any social norms and is not tested in environments with faster-moving objects or multiple dynamic agents.

\section{Conclusions \& Future Work}
\label{sec:conclusion}
In this work, we presented a perceptive Forward Dynamics Model framework for deployment in challenging local planning tasks.
Trained with a mix of simulated and real-world data, the FDM captures the complex dynamics of a quadrupedal robot and enables zero-shot adjustments of the planning objective.
The presented network decreases position errors by, on average, 41.28\% compared to baseline methods and estimated failures with an accuracy of at least 89.20\%.
Moreover, our FDM integrated into an MPPI planner with simplified rewards achieves on average 81\% goal success rate in complex environments. 

For future work, we will explore adaptive timesteps for applied commands and extend the range of addressed environments. Additionally, we aim to transition to RGB input for a richer environmental representation. We also plan to integrate the proposed FDM into an ensemble learning framework to assess uncertainty and use it as an additional planning parameter. Furthermore, this work can serve as a step toward improving the fidelity of physics simulators in challenging environments.

\section*{Acknowledgments}
The authors thank Fan Yang for their support and scientific discussions. This work is supported by the Swiss National Science Foundation (SNSF) as part of project No.227617, ETH RobotX research grant funded through the ETH Zurich Foundation, the European Union's Horizon research and innovation program under grant agreement No 101070596, No 101070405, and No 852044, and an ETH Zurich Research Grant No. 21-1 ETH-27. Jonas Frey is supported by the Max Planck ETH Center for Learning Systems.

\bibliographystyle{unsrtnat}
\bibliography{bibliography/references}

\newpage
\onecolumn

\appendix
\subsection{Nomenclature}

\begin{table}[h!]
    \centering
    \renewcommand{\arraystretch}{1.2}
    \resizebox{\textwidth}{!}{
    \begin{tabular}{p{2cm} p{6cm} p{2cm} p{6cm}}
        \toprule
        \textbf{Symbol} & \textbf{Description} & \textbf{Symbol} & \textbf{Description} \\
        \midrule
        $a$ & Action & $\tilde{a}$ & Action correction predicted by FDM \\
        $\hat{a}$ & Followed action on the robot & $b$ & Robot joint count \\
        $f$ & FDM function & $g$ & Goal pose \\
        $h$ & Height scan & $k$ & MPPI iteration count \\
        $m$ & Proprioceptive measurements & $n$ & FDM number of prediction steps \\
        $o$ & Observations & $p$ & Pose \\
        $q$ & Set of neighbors for MPPI obstacle cost & $r$ & Risk \\
        $s$ & State & $t$ & Time \\
        $u$ & Height-map width & $v$ & Height-map length \\
        $w$ & Trajectory weight assigned for MPPI update & $z$ & Observation probability \\
        \midrule
        $\mathcal{L}$ & FDM Loss & $\mathcal{S}$ & Set of states \\
        $\mathcal{T}$ & Transition likelihood between states & $\mathcal{O}$ & Set of observations \\
        $\mathcal{Z}$ & Observation probability & $\mathcal{C}$ & Set of MPPI candidates \\
        $\mathcal{R}$ & MPPI Reward functions & $\mathcal{U}$ & Uniform distribution \\
        $\mathcal{N}$ & Normal Distribution &  &  \\        
        \midrule
        $\beta$ & Time correlation factor for action sampling & $\sigma$ & Standard deviation for action sampling \\        
        $\theta$ & FDM network weights & $\lambda$ & MPPI reward term weights \\
        $\epsilon$ & FDM loss term weights & $\delta_{risk}$ & Threshold for risky trajectory \\
        $\delta_{pose}$ & Threshold to apply pull factor in pose reward &  \\
        \midrule
        $\Delta t_h$ & Time-step of the history information with frequency $1 / \Delta t_h$ & $\Delta t_p$ & Time-step of the forward predictions with frequency $1 / \Delta t_p$ \\
        \bottomrule
    \end{tabular}
    }
    \caption{Nomenclature used in this work.}
    \label{tab:nomenclature}
\end{table}

\subsection{Time-Correlated action sampling}
\label{subsec:time-correlated-action-sampling}

To model the time-correlated action given by the MPPI planner and capture a broad distribution of action sequences, we deploy linear time-correlated command sampling (Eq.~\ref{eq:linear}) and normal time-correlated command sampling (Eq.~\ref{eq:normal}). All sequences are clipped to the minimum and maximum values for the different velocities.

\vspace{0.5cm}

\noindent Linear time-correlated command sampling with time correlation factor $\beta$ and the normal distribution $\mathcal{U}$:
\begin{equation}
    \beta \sim \mathcal{U}(\beta_{\text{min}}, 1), \quad a_{\text{rand}} \sim \mathcal{U}(a_{\text{min}}, a_{\text{max}})
\end{equation}
\begin{equation}
    a_{t+1} = \beta \cdot a_t + (1 - \beta) \cdot a_{\text{rand}}, \quad \forall t \in \{0, \ldots, n-1\}
\label{eq:linear}
\end{equation}

\noindent Normal time-correlated command sampling with the standard deviation $\sigma$ and the normal distribution $\mathcal{N}$:
\begin{equation}
\sigma \sim \mathcal{U}(0, \sigma_{\text{max}})
\end{equation}
\begin{equation}
a_{t+1} \sim \mathcal{N}(a_t, \sigma), \quad \forall t \in \{0, \ldots, n-1\}
\label{eq:normal}
\end{equation}

\subsection{Detailed Model Structure}

The model has been designed for efficient deployment on a mobile robot with limited computing. Consequently, the architecture is restricted to CNN and MLP layers, compromising 1.16 million parameters overall. The detailed structure, as an example for the quadrupedal robot ANYmal~\cite{hutter2016anymal}, is as follows:

\begin{lstlisting}[frame=single]
    FDMModelMultiStep(
  (state_obs_proprioceptive_encoder): GRU(
    66, 64, num_layers=2, batch_first=True, dropout=0.2
  )
  (obs_exteroceptive_encoder): CNN(
    (architecture): Sequential(
      (0): Conv2d(1, 32, kernel_size=(7, 7), stride=(1, 1))
      (1): LeakyReLU(negative_slope=0.01)
      (2): MaxPool2d(kernel_size=3, stride=2, padding=1, dilation=1, ceil_mode=False)
      (3): Conv2d(32, 64, kernel_size=(3, 3), stride=(2, 2))
      (4): LeakyReLU(negative_slope=0.01)
      (5): Conv2d(64, 128, kernel_size=(3, 3), stride=(2, 2))
      (6): LeakyReLU(negative_slope=0.01)
      (7): Conv2d(128, 256, kernel_size=(3, 3), stride=(2, 2))
      (8): LeakyReLU(negative_slope=0.01)
    )
  )
  (action_encoder): MLP(
    (architecture): Sequential(
      (0): Linear(in_features=3, out_features=16, bias=True)
      (1): LeakyReLU(negative_slope=0.01)
      (2): Dropout(p=0.2, inplace=False)
    )
  )
  (recurrence): GRU(596, 128, num_layers=2, batch_first=True, dropout=0.2)
  (state_predictor): MLP(
    (architecture): Sequential(
      (0): Linear(in_features=1280, out_features=128, bias=True)
      (1): LeakyReLU(negative_slope=0.01)
      (2): Linear(in_features=128, out_features=64, bias=True)
      (3): LeakyReLU(negative_slope=0.01)
      (4): Dropout(p=0.2, inplace=False)
      (5): Linear(in_features=64, out_features=30, bias=True)
    )
  )
  (collision_predictor): MLP(
    (architecture): Sequential(
      (0): Linear(in_features=1280, out_features=64, bias=True)
      (1): LeakyReLU(negative_slope=0.01)
      (2): Linear(in_features=64, out_features=10, bias=True)
    )
  )
  (sigmoid): Sigmoid()
)
\end{lstlisting}

\subsection{Design Ablations}

We conducted ablation studies to assess the contributions of past states, proprioceptive inputs, and exteroceptive height scans across multiple environments. Detailed results are presented in Tab.~\ref{table:model_ablations}.
Incorporating past states enhances failure estimation across all environments, increasing the F1 score by up to 0.06. However, it does not impact pose accuracy. In contrast, removing proprioceptive inputs significantly degrades both performances, leading to higher position errors and lower F1 scores. The exteroceptive height scan proves critical for the perceptive navigation task. Excluding it results in an average 110\% increase in position error and a 0.13 decrease in the F1 score. 
Further, we evaluated the impact of removing the failure risk reward from the planner, with results summarized in Tab.~\ref{tab:ablation_study}. When omitted, planning success decreased by 2.5\% in 2D environments and 4.6\% in 3D environments. Although the mean path length remained nearly constant, the introduction of safety constraints led to longer mean path times. These findings indicate that failure risk estimation contributes to higher success rates, though it results in more cautious and slower planning.

\begin{table}[h!]
\centering
\begin{tabular}{>{\centering\arraybackslash}p{0.3cm}lccccc}
    \hline
    \textbf{Env.} & \textbf{Method Variation} & \textbf{Pos. Offset} & \textbf{Precision} & \textbf{Recall} & \textbf{Accuracy} & \textbf{F1 Score} \\
    \hline
    \multirow{4}{*}{\rotatebox{90}{Plane}}
    & Ours & \textbf{0.14 $\pm$ 0.16} & \textcolor{gray}{\textbf{41.69}} & \textcolor{gray}{35.23} & \textcolor{gray}{98.88} & \textcolor{gray}{\textbf{0.37}} \\
    & Ours W/o State Obs. & 0.14 $\pm$ 0.17 & \textcolor{gray}{30.12} & \textcolor{gray}{26.89} & \textcolor{gray}{\textbf{99.08}} & \textcolor{gray}{0.28} \\
    & Ours W/o Proprio. Obs. & 1.22 $\pm$ 0.81 & \textcolor{gray}{2.62} & \textcolor{gray}{\textbf{81.02}} & \textcolor{gray}{83.38} & \textcolor{gray}{0.05} \\
    & Ours W/o Height Scan & 0.44 $\pm$ 0.40 & \textcolor{gray}{5.92} & \textcolor{gray}{63.51} & \textcolor{gray}{92.82} & \textcolor{gray}{0.10} \\
    \hline
    \multirow{4}{*}{\rotatebox{90}{2D}}
    & Ours & 0.23 $\pm$ 0.28 & \textbf{93.7} & 90.14 & \textbf{91.43} & \textbf{0.92} \\
    & Ours W/o State Obs. & \textbf{0.22 $\pm$ 0.27} & 89.38 & 89.23 & 90.93 & 0.89 \\
    & Ours W/o Proprio. Obs. & 0.25 $\pm$ 0.32 & 86.05 & \textbf{91.00} & 89.77 & 0.88 \\
    & Ours W/o Height Scan & 0.48 $\pm$ 0.51 & 89.65 & 68.89 & 82.75 & 0.78 \\
    \hline
    \multirow{4}{*}{\rotatebox{90}{2D-3D}}
    & Ours & \textbf{0.26 $\pm$ 0.32} & \textbf{93.61} & \textbf{91.56} & \textbf{93.24} & \textbf{0.93} \\
    & Ours W/o State Obs. & 0.26 $\pm$ 0.32 & 88.18 & 85.69 & 92.07 & 0.87 \\
    & Ours W/o Proprio. Obs. & 0.32 $\pm$ 0.35 & 83.64 & 87.89 & 90.73 & 0.86 \\
    & Ours W/o Height Scan & 0.43 $\pm$ 0.43 & 87.35 & 72.36 & 87.39 & 0.79 \\
    \hline
    \multirow{4}{*}{\rotatebox{90}{3D}}
    & Ours & \textbf{0.27 $\pm$ 0.32} & \textbf{90.28} & \textbf{85.9} & \textbf{93.14} & \textbf{0.88} \\
    & Ours W/o State Obs. & 0.27 $\pm$ 0.32 & 89.41 & 85.12 & 92.70 & 0.87 \\
    & Ours W/o Proprio. Obs. & 0.28 $\pm$ 0.32 & 85.34 & 87.09 & 92.04 & 0.86 \\
    & Ours W/o Height Scan & 0.41 $\pm$ 0.41 & 88.94 & 72.61 & 88.89 & 0.80 \\
    \hline
\end{tabular}
\caption{Ablation studies evaluating the impact of removing specific input modalities across multiple environments. Omitting past state information reduces the F1 score across all environments. Removal of proprioceptive inputs or the height scan results in substantial declines in both pose accuracy and failure prediction, underscoring the importance of these components.}
\label{table:model_ablations}
\end{table}

\begin{table}[h!]
    \centering
    \begin{tabular}{clccc}
        \hline
        \textbf{Env.} & \textbf{Method} & \textbf{Success (\%)} & \textbf{MPL (m)} & \textbf{MPT (s)} \\
        \hline
        \multirow{2}{*}{2D}
        & Ours & \textbf{88.33} & \textbf{4.28} & 9.23 \\
        & Ours (w/o Failure Estimation) & 85.86 & 4.50 & \textbf{7.73} \\
        \hline
        \multirow{2}{*}{3D}
        & Ours & \textbf{73.75} & \textbf{3.93} & 8.68 \\
        & Ours (w/o Failure Estimation) & 69.17 & 3.96 & \textbf{5.74} \\
        \hline
    \end{tabular}
    \caption{Influence of the risk term when planning in $2D$ and $3D$ environments. By including the risk reward term, the success rate improves in both cases. While the mean path length (MPT) is approximately equal, the additional safety requirement leads to longer mean path times (MPT).}
    \label{tab:ablation_study}
\end{table}

\FloatBarrier
\subsection{Terrain Details}
\label{subsec:terrain_details}

\begin{figure*}[ht!]
    \centering
    \includegraphics[width=1.0\textwidth]{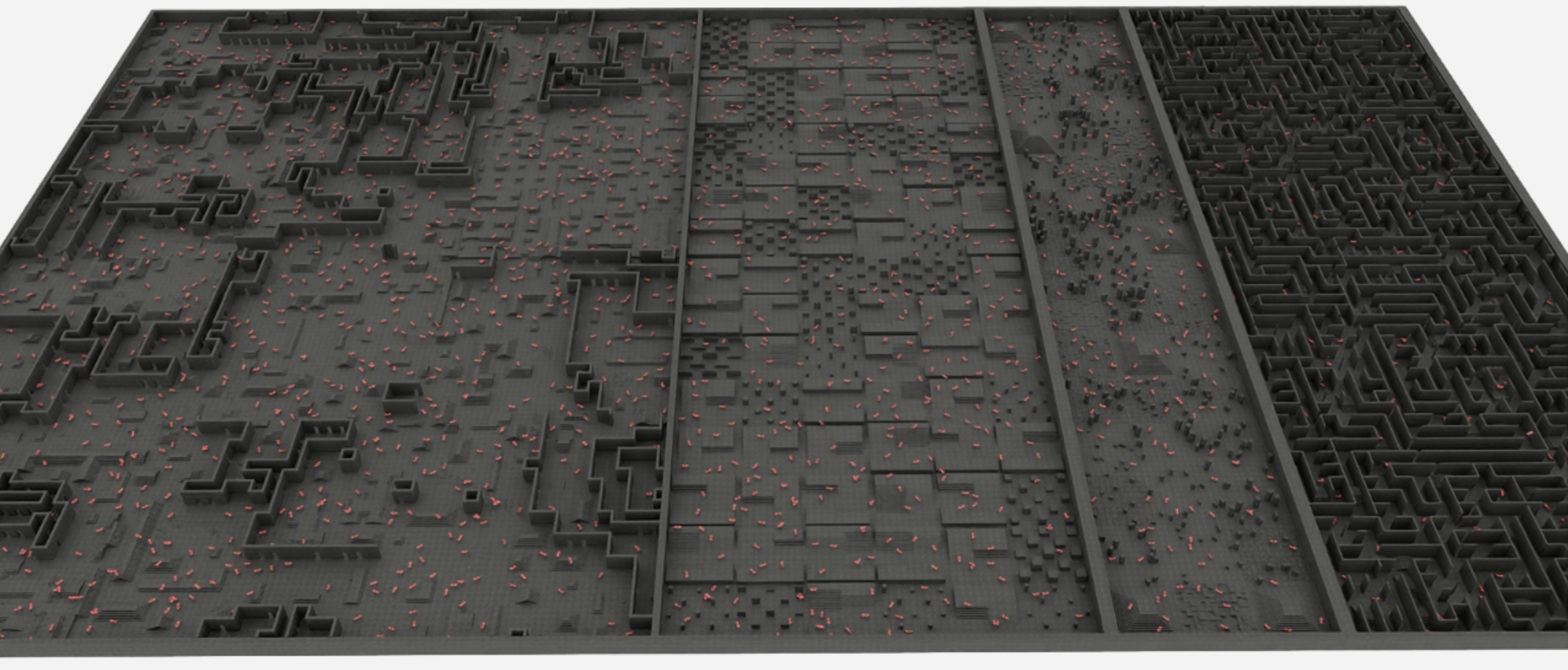}
    \caption{The simulation training environment consists of four distinct segments. The first segment features a randomized mix of stairs, ramps, walls, and rough surfaces. The second contains obstacles arranged in structured patterns. The third includes pillars of varying dimensions placed on different surfaces. The final segment is a randomly generated maze.}
    \label{fig:training-env}
\end{figure*}

The simulation terrain consists of four distinct segments, as displayed in Fig.~\ref{fig:training-env}. The largest segment~(44\%) combines various tiles (stairs, ramps, flat and rough ground, walls, doors, etc.), resulting in a highly randomized terrain. Adjacent to this is a structured segment~(22\%), featuring stairs, ramps, and obstacles like walls, boxes, and pillars in structured patterns with randomized dimensions. The third segment~(12\%) comprises randomly placed pillars on different surfaces with variations in the pillar dimensions. Lastly, a random maze terrain is used with additional stairs between the walls~(22\%). While we did not apply domain randomization, the locomotion policies used were trained with randomization on friction, body masses, inertia, and random disturbances.

\subsection{Expanded Planning Experiments Results}
\label{app:planning_exp}

\begin{figure}[ht]
    \centering
    \begin{subfigure}[b]{0.45\textwidth}
        \includegraphics[width=\textwidth]{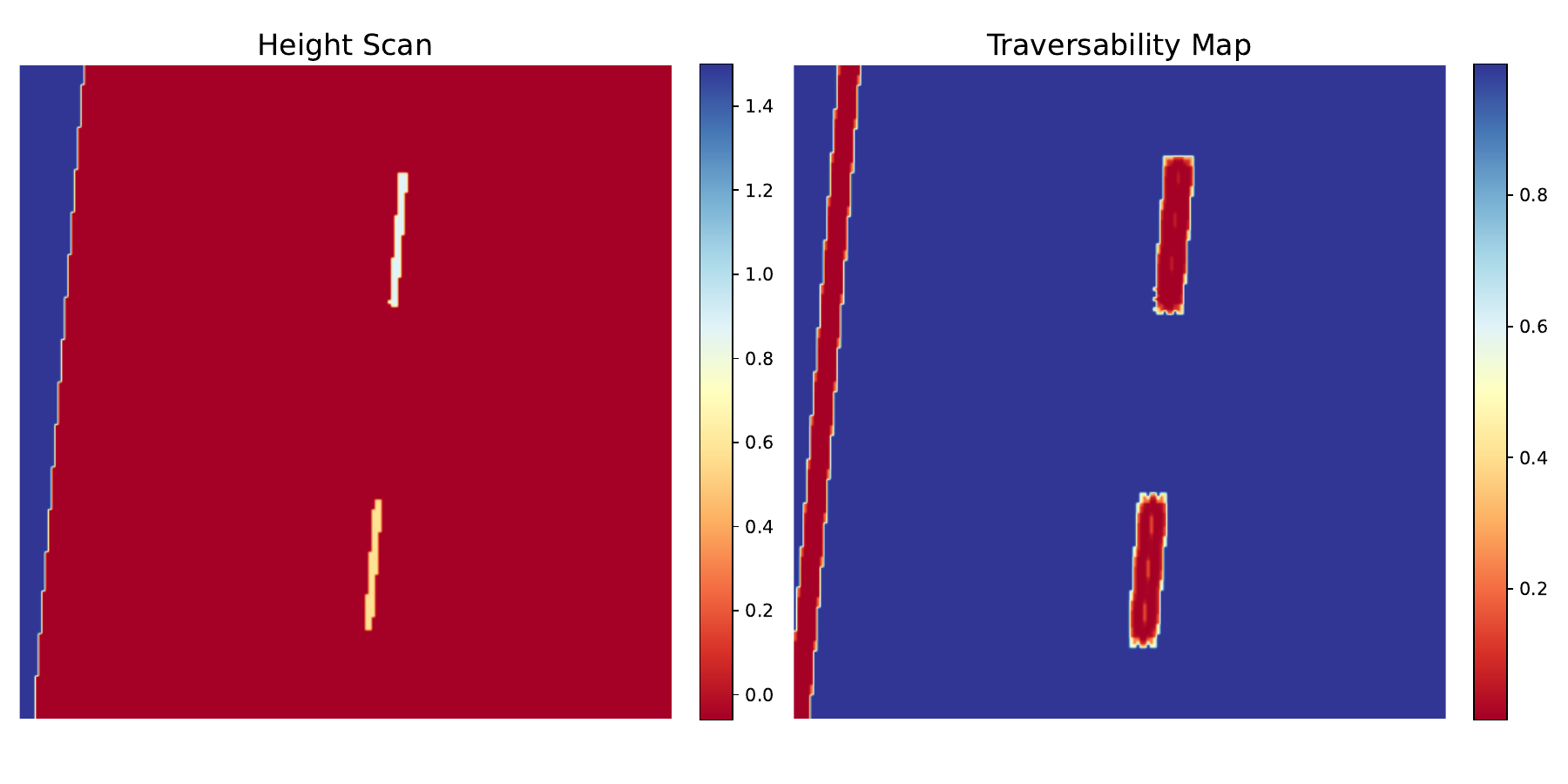}
    \end{subfigure}
    \hfill
    \begin{subfigure}[b]{0.45\textwidth}
        \includegraphics[width=\textwidth]{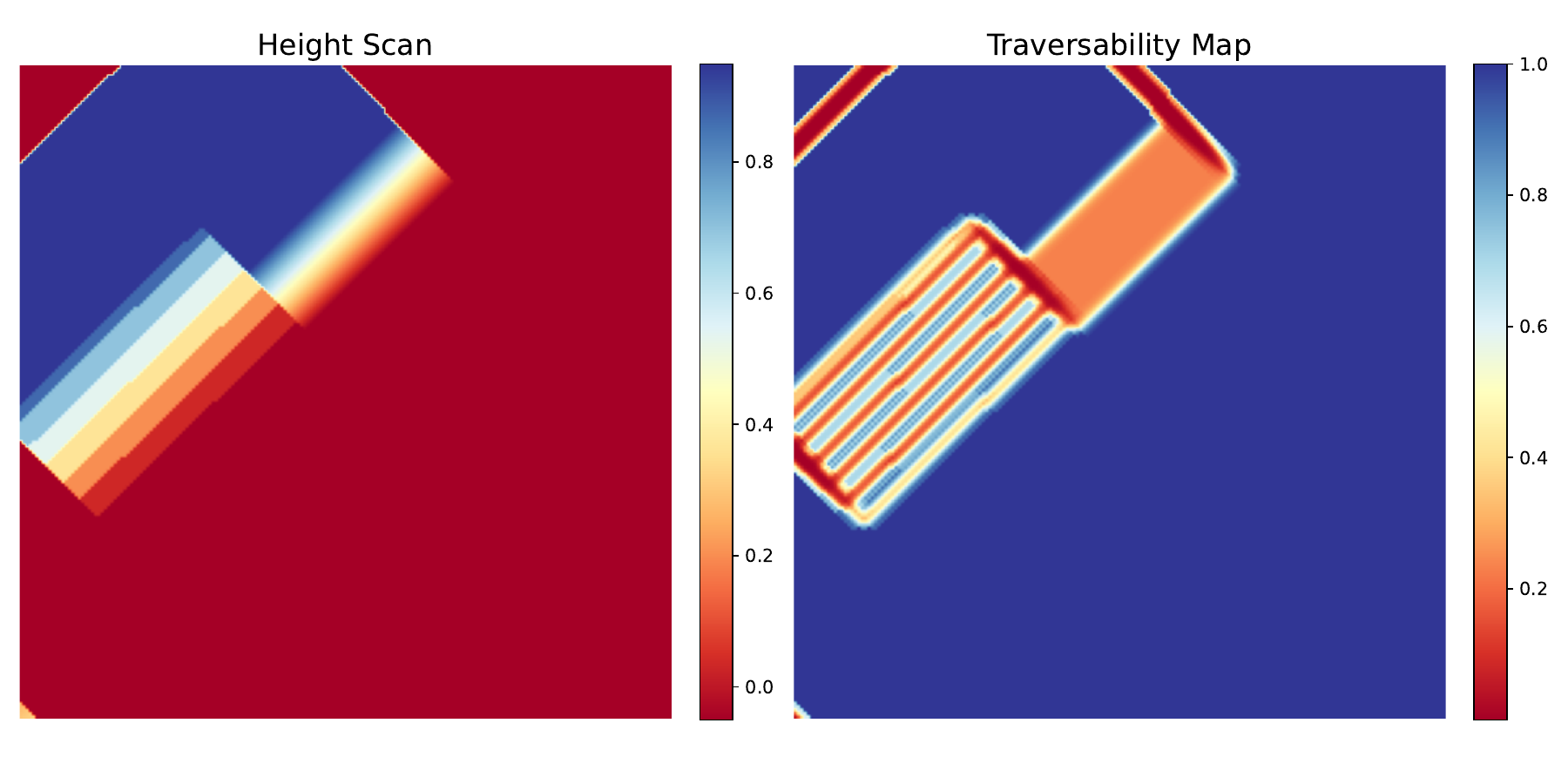}
    \end{subfigure}

    \begin{subfigure}[b]{0.45\textwidth}
        \includegraphics[width=\textwidth]{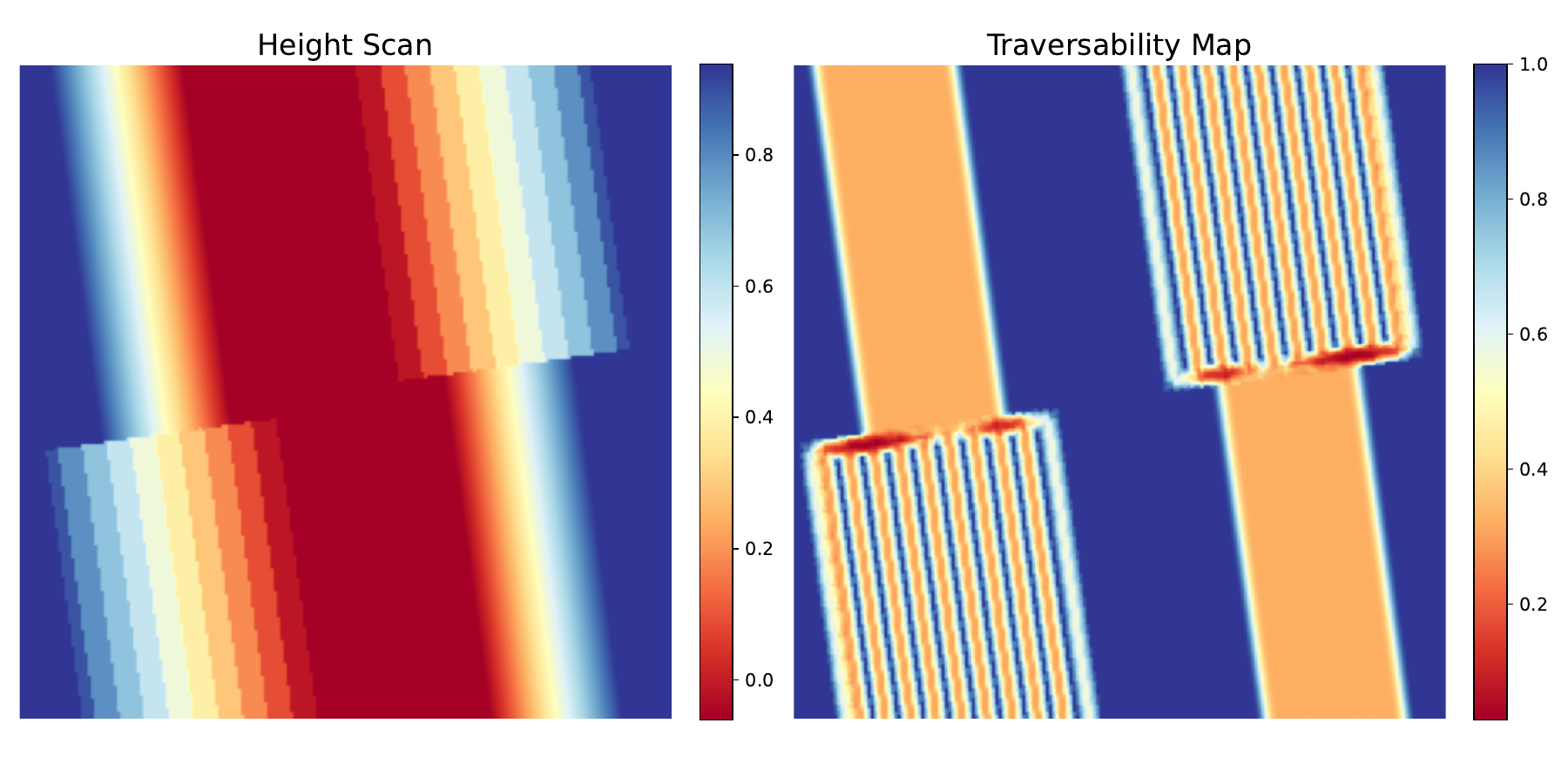}
    \end{subfigure}
    \hfill
    \begin{subfigure}[b]{0.45\textwidth}
        \includegraphics[width=\textwidth]{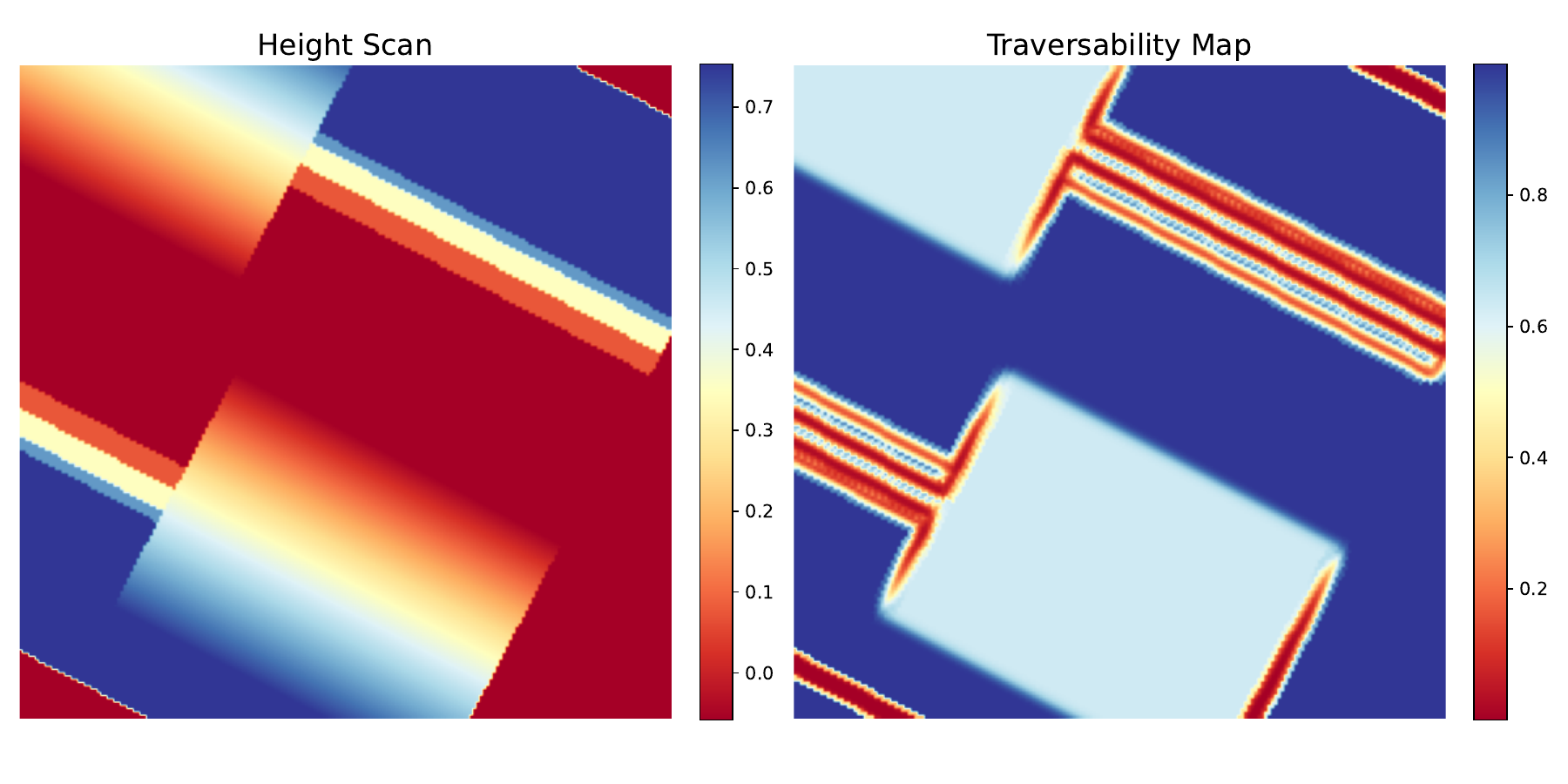}
    \end{subfigure}

    \caption{Combined visualization of the height scan and traversability estimates generated by the heuristics-based method of~\cite{wellhausen2023artplanner} for four environments. It is visible that stairs often have very low traversability scores, even if they should be traversable. Nevertheless, the scores are in the same range as the ramps, which are too steep to traverse, making tuning for arbitrary environments challenging.}
    \label{fig:heuristic-trav-estimate}
\end{figure}

The MPPI planner was tuned specifically for the baseline methods of~\citet{kim2022learning} and~\citet{wellhausen2021rough} to achieve the results reported in Tab.~\ref{tab:planning_performance}. For the former, the population size was reduced by half (to 256), and the weight of the risk reward term was decreased by a factor of 10. These adjustments were necessary due to the method's conservative nature, which tends to classify paths near obstacles as high-risk, hindering its ability to plan effectively in narrow environments. Additionally, the time-out threshold was increased to allow more extensive exploration.
For the heuristic-based method, the risk reward weight remained unchanged from our setup. Only the neighboring filter parameter, introduced in Sec.~\ref{subsec:planning_loss}, was increased from three to four to ensure a greater safety margin around obstacles, compensating for the less accurate constant-velocity assumption.
Fig.~\ref{fig:heuristic-trav-estimate} visualizes traversability estimates from the heuristics-based method. It illustrates the tuning challenges in complex environments, particularly in differentiating between stairs and steep ramps.

\subsection{Expanded Dynamics Experiments Results}
\label{app:expanded_exp}

\begin{figure}[ht!]
    \centering
    \includegraphics[width=1.0\columnwidth]{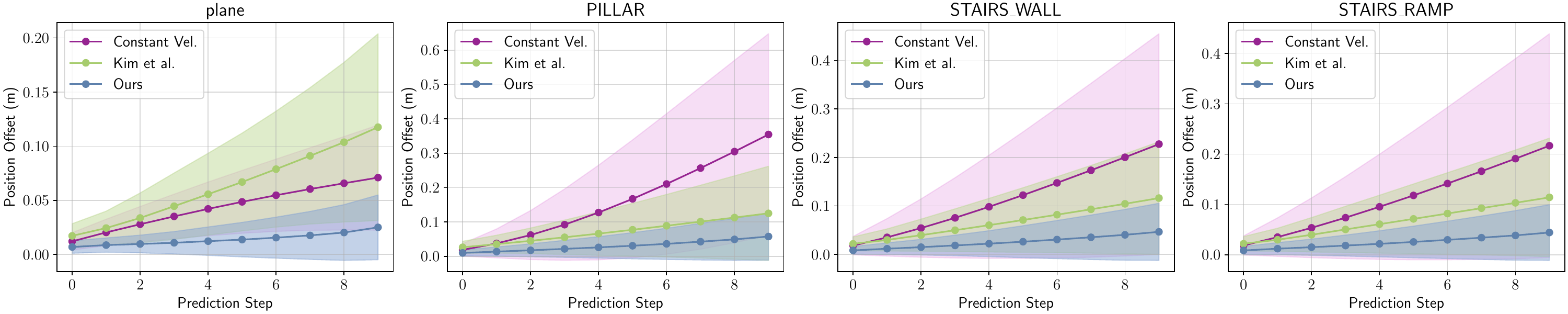}
    \caption{Comparison of the position error over the prediction steps between the presented method \textcolor{ours}{\rule{1.5ex}{1.5ex}}, the perceptive FDM by~\citet{kim2022learning} \textcolor{baseline}{\rule{1.5ex}{1.5ex}}, and the constant velocity model \textcolor{constant_vel}{\rule{1.5ex}{1.5ex}} for the different environments. Our FDM demonstrates the highest accuracy and smallest standard deviation across all environments.}
    \label{fig:performance-over-time-non-avg}
\end{figure}

\begin{figure*}[ht!]
    \centering
    \includegraphics[width=0.9\textwidth]{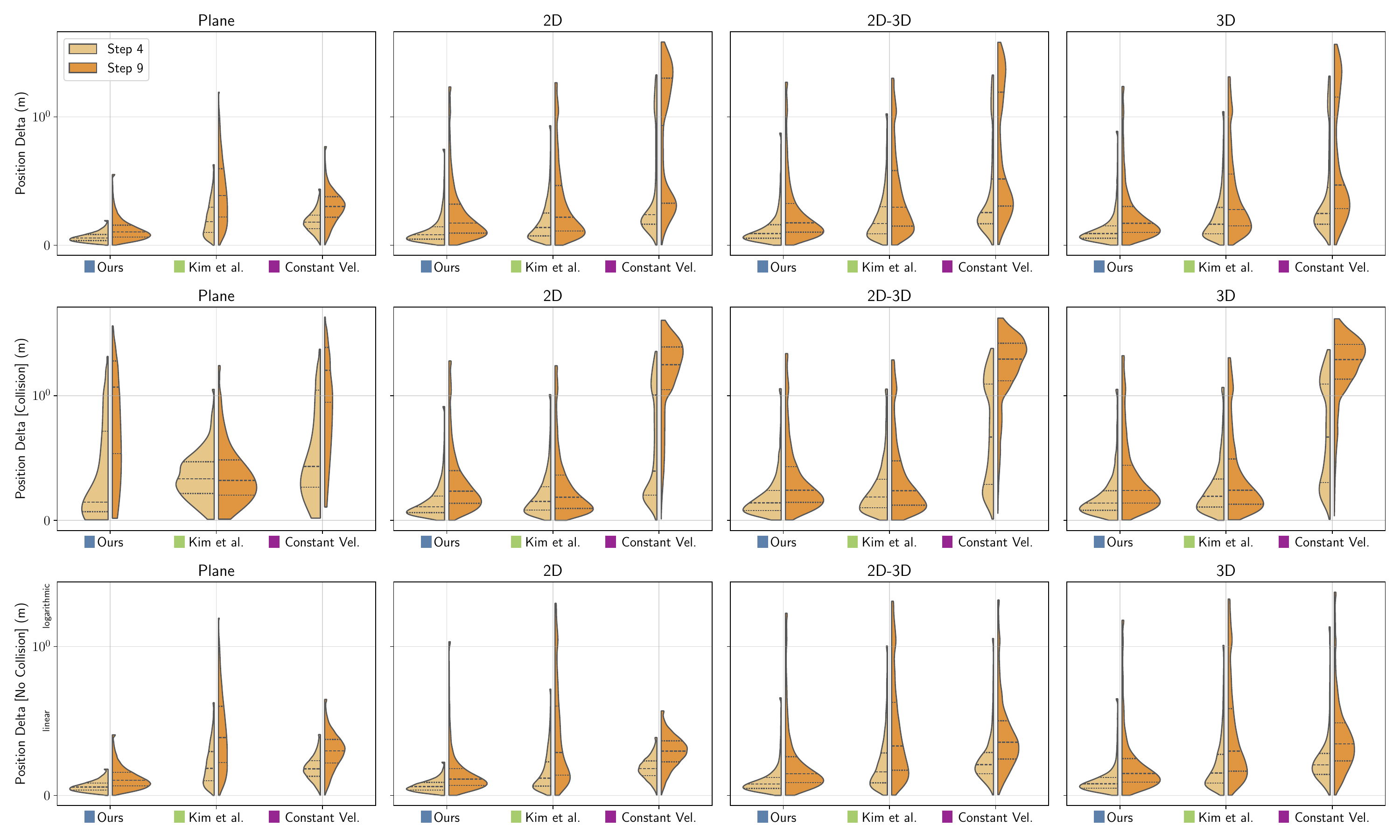}
    \caption{Comparison of position error at the final prediction step across different environments for the presented FDM \textcolor{ours}{\rule{1.5ex}{1.5ex}}, the perceptive FDM by~\citet{kim2022learning} \textcolor{baseline}{\rule{1.5ex}{1.5ex}}, and the constant velocity model \textcolor{constant_vel}{\rule{1.5ex}{1.5ex}}. For each environment, 50k samples are evaluated, and the error is shown up to the 95\% quantile to minimize the impact of outliers. Errors are split between collision and non-collision samples. In the $2D-3D$ and $3D$ environments, the perceptive baseline struggles to differentiate obstacles, resulting in higher errors for non-collision cases, as it mistakenly predicts collisions. In contrast, the proposed method shows significantly lower error rates in these scenarios. Failures in the planar case are non-deterministic and likely due to simulation instabilities that the models cannot predict, leading to larger error scores.}
    \label{fig:performance-final-step-coll-split}
\end{figure*}

\begin{figure}[ht!]
    \centering
    \includegraphics[width=0.9\columnwidth]{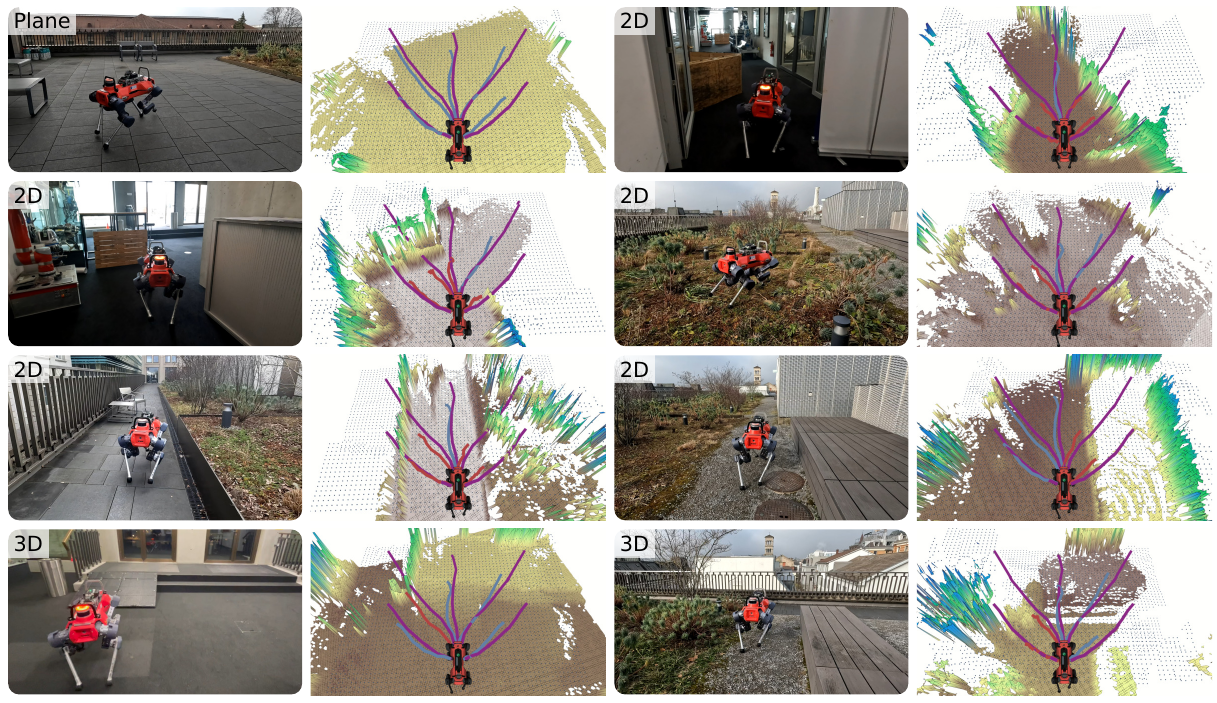}
    \caption{Additional real-world demonstration of environment- and platform-aware state predictions using the presented FDM in comparison to constant velocity estimation \textcolor{constant_vel}{\rule{1.5ex}{1.5ex}} for the same action sequences across multiple scenarios. Collision-free predictions of our method are displayed in \textcolor{ours}{\rule{1.5ex}{1.5ex}}, in collision ones in \textcolor{collision}{\rule{1.5ex}{1.5ex}}.}
    \label{fig:test_perceptive_fdm_expanded}
\end{figure}

This subsection presents an expanded set of experimental results evaluating the performance of the proposed FDM. 
Fig.~\ref{fig:performance-over-time-non-avg} showcases how the position error evolves over the prediction horizon across the different environments, where our method consistently achieves higher accuracy and lower variance than baseline models. 
Fig.~\ref{fig:performance-final-step-coll-split} compares the position error across different environments at the final prediction step. It expands Fig.~\ref{fig:performance-final-step} with an additional split between collision and non-collision scenarios. 
Additional real-world demonstrations of the environment- and platform-aware predictions are provided in Fig.~\ref{fig:test_perceptive_fdm_expanded}.

\FloatBarrier
\subsection{Adaptation for new Platforms and Parameter Sensitivity}
\label{app:adaptation}
Adapting our FDM to a new robot platform requires updating the simulation model, locomotion policy, and input layers to reflect platform-specific features such as additional joints or proprioceptive information. 
Moreover, command sampling might be adjusted for each platform. 
We assume network architecture and learning hyperparameters can remain unchanged. 
The embodiment-specific predictions are a key benefit of our method, tailoring the FDM to the platform’s hardware and software. 

Within our tests, we found that our MPPI parameters are more sensitive compared to the learning setup. Specifically, complex scenarios require carefully tuned parameters that facilitate path variations to overcome local minima. Key tuning parameters include noise magnitude, time correlation, and reward scaling. 
Additionally, the sampling space, which determines which actions the robot is allowed to perform, e.g., omnidirectional motion or walking backwards, is a key design decision that allows for reduced cost term tuning. 
Experimentally, we found that higher noise values are generally preferred, as MPPI prioritizes selecting the highest-reward trajectory across subsequent iterations.

\end{document}